\documentclass[lettersize,journal]{IEEEtran}
\usepackage{array}
\usepackage[caption=false,font=normalsize,labelfont=sf,textfont=sf]{subfig}
\usepackage{stfloats}
\usepackage{verbatim}
\usepackage{cite}
\usepackage{orcidlink}
\usepackage{url}
\usepackage{graphicx}
\usepackage{multirow}
\usepackage{amsmath,amssymb,amsfonts}
\usepackage{amsthm}
\usepackage{mathrsfs}
\usepackage{xcolor}
\usepackage{textcomp}
\usepackage{manyfoot}
\usepackage{booktabs}
\usepackage{algorithmicx}
\usepackage{algpseudocode}
\usepackage{listings}

\usepackage[small]{caption}
\usepackage{subfig}
\usepackage{pifont}
\usepackage{enumitem}
\usepackage{soul, color}
\usepackage[ruled, linesnumbered]{algorithm2e}
\usepackage{colortbl}
\usepackage{makecell}

\begin{document}

\title{Model Inversion Attacks Through Target-Specific Conditional Diffusion Models}

\author{Ouxiang Li, Yanbin Hao, Zhicai Wang, Bin Zhu, Shuo Wang, Zaixi Zhang, Fuli Feng

\thanks{Ouxiang Li, Zhicai Wang, and Fuli Feng are with School of Artificial Intelligence and Data Science, University of Science and Technology of China, Hefei, China (e-mail: lioox@mail.ustc.edu.cn, wangzhic@mail.ustc.edu.cn, fulifeng93@gmail.com).}
\thanks{Yanbin Hao is with School of Computer Science and Information Engineering, Hefei University of Technology, Hefei, 230601, China (e-mail: haoyanbin@hotmail.com).}
\thanks{Bin Zhu is with School of Computing and Information Systems, Singapore Management University, Singapore (e-mail: binzhu@smu.edu.sg).}
\thanks{Shuo Wang is with School of Information Science and Technology, University of Science and Technology of China, Hefei, China (e-mail: shuowang.edu@gmail.com).}
\thanks{Zaixi Zhang is with Department of Computer Science and Engineering, University of Science and Technology of China, Hefei, China (e-mail: zaixi@mail.ustc.edu.cn).}
}

\markboth{Journal of \LaTeX\ Class Files,~Vol.~14, No.~8, August~2021}%
{Shell \MakeLowercase{\textit{et al.}}: A Sample Article Using IEEEtran.cls for IEEE Journals}


\maketitle

\begin{abstract}
Model inversion attacks (MIAs) aim to reconstruct private images from a target classifier's training set, thereby raising privacy concerns in AI applications. Previous GAN-based MIAs tend to suffer from inferior generative fidelity due to GAN's inherent flaws and biased optimization within latent space. To alleviate these issues, leveraging on diffusion models' remarkable synthesis capabilities, we propose Diffusion-based Model Inversion (Diff-MI) attacks. Specifically, we introduce a novel target-specific conditional diffusion model (CDM) to purposely approximate target classifier's private distribution and achieve superior accuracy-fidelity balance. Our method involves a two-step learning paradigm. Step-1 incorporates the target classifier into the entire CDM learning under a pretrain-then-finetune fashion, with creating pseudo-labels as model conditions in pretraining and adjusting specified layers with image predictions in fine-tuning. Step-2 presents an iterative image reconstruction method, further enhancing the attack performance through a combination of diffusion priors and target knowledge. Additionally, we propose an improved max-margin loss that replaces the hard max with top-k maxes, fully leveraging feature information and soft labels from the target classifier. Extensive experiments demonstrate that Diff-MI significantly improves generative fidelity with an average decrease of 20\% in FID while maintaining competitive attack accuracy compared to state-of-the-art methods across various datasets and models. Our code is available at: \url{https://github.com/Ouxiang-Li/Diff-MI}.
\end{abstract}


\section{Introduction} \label{sec:1}

Deep Neural Networks (DNNs) are revolutionizing various fields such as computer vision, autonomous driving, and healthcare. However, these advancements have also raised significant concerns regarding privacy attacks on DNNs. One such category is Model Inversion Attacks (MIAs), which aim to reconstruct private training samples by exploiting a ``target classifier". In this paper, we focus on the white-box MIA to reconstruct user images, in which case the attackers are assumed to have full access to the target classifier.

Previous works \cite{zhang2020secret,chen2021knowledge,yuan2023pseudo} in the white-box setting have succeeded in reconstructing private images disclosing personal information by using generative adversarial networks (GANs) \cite{goodfellow2014generative}. Firstly, they train the GAN on a public dataset to learn an image prior that only shares structural similarity with the private dataset (i.e., without any interclass overlap). Then leveraging the full accessibility of the target classifier, they continuously optimize latent variables through iterations until they could reconstruct images with the highest confidence associated with specific target labels in the target classifier. In a word, their works can be summarized as a single-objective optimization problem for the latent variables $\mathbf{z}$, which are initially sampled from a Standard Gaussian distribution $\mathcal{N}(\mathbf{0}, \mathbf{I})$, by minimizing the classification loss between reconstructed images and target labels.

\begin{figure*}[t]
    \centering
    \includegraphics[width=1.00\textwidth]{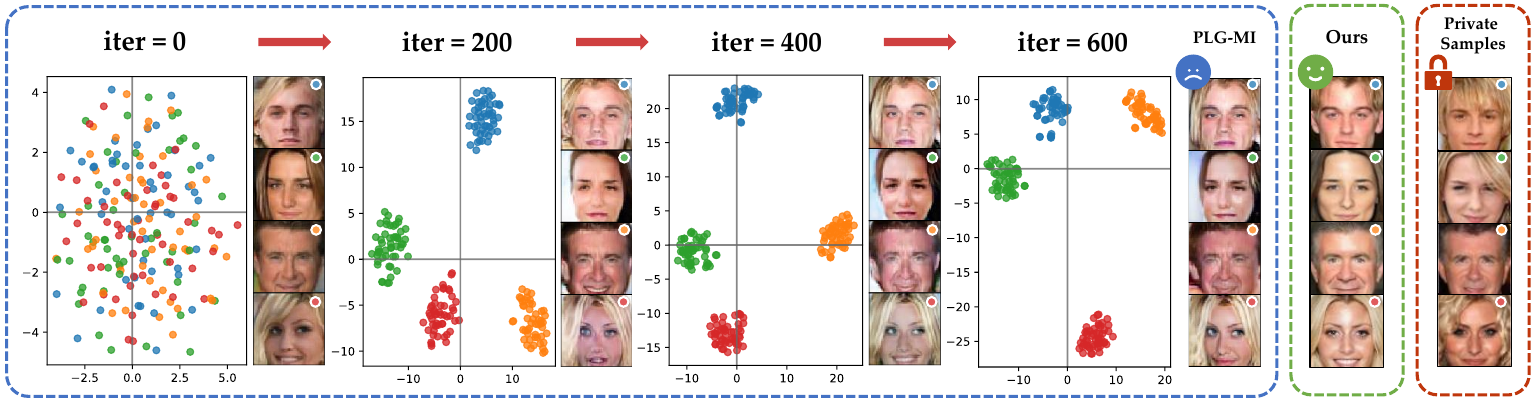}
    \caption{
    \textbf{Fidelity degradation.} We randomly select 4 target classes and visualize their distributional variation in GAN's latent space along with corresponding generated images, depicted with different colors (MIA Method = PLG-MI \protect\cite{yuan2023pseudo}, Private Dataset = CelebA \protect\cite{liu2015deep}, Public Dataset = CelebA, Target Classifier = Face.evoLVe \protect\cite{cheng2017know}). Through iterations, we observe that the latent variables, initially sampled from $\mathcal{N}(\mathbf{0}, \mathbf{I})$, tend to cluster together and constitute a new distribution with a deviation from $\mathcal{N}(\mathbf{0}, \mathbf{I})$. Meanwhile, reconstructed images gradually deteriorate visually during inversion because of the widening gap between the optimized latent distribution and the prior distribution. As a result, PLG-MI sacrifices generative fidelity for attack accuracy, which is also present in other GAN-based methods \protect\cite{zhang2020secret,chen2021knowledge}.
    }
    \label{fig:1}
\end{figure*}

Successful MIAs should balance both attack accuracy and fidelity. However, these works using GANs as image priors commonly suffer from inferior generative fidelity, which can be attributed to the following two reasons. On the one hand, GANs are difficult to train and prone to collapse without carefully tuned hyper-parameters and regularizers \cite{brock2018large,miyato2018spectral,dhariwal2021diffusion}, inherently falling short in generative quality. On the other hand, since GANs can be viewed as a mapping from a known distribution (e.g., Standard Gaussian distribution) to a complex image distribution \cite{goodfellow2014generative}, such an optimization strategy can potentially distort the prior distribution in the latent space, leading to \textbf{fidelity degradation} (see Fig.~\ref{fig:1}) in pursuit of higher attack accuracy. Additionally, previous works have improved the standard cross-entropy (CE) loss in MIAs for Poincaré loss \cite{struppek2022plug} and max-margin loss \cite{yuan2023pseudo}, whereas the feature information and soft labels provided by the target classifier still remains underutilized.

% In contrast with GANs, diffusion models (DMs) \cite{ho2020denoising} are likelihood-based generative models which have recently been shown to produce impressive high-quality images \cite{dhariwal2021diffusion,ho2022classifier,rombach2022high}. 
To address the above limitations, we propose a two-step Diffusion-based Model Inversion (Diff-MI) attack, leveraging on diffusion models (DMs)' remarkable synthesis capabilities \cite{dhariwal2021diffusion,ho2022classifier,rombach2022high}. Specifically,  in step-1, we devise a target-specific conditional diffusion model (CDM) to distill the target classifier's knowledge and approximate its private distribution. Particularly, we first pretrain a CDM with public images along with their pseudo-labels produced by the target classifier as conditions. To fully distill the target classifier's white-box knowledge (e.g., gradients), we then fine-tune a small subset of the pretrained CDM through its sampling process with image predictions by the target classifier, achieving superior accuracy-fidelity balance. In step-2, we introduce an iterative image reconstruction method for diffusion-based MIAs to mitigate the fidelity degradation problem. It turns the single-objective optimization problem (i.e., minimizing the classification loss) into a combinatorial optimization problem by involving both learned CDM prior and target classifier constraint, further enhancing the attack performance. Additionally, we propose an improved max-margin loss replacing the hard max with top-k maxes along with a p-reg loss regularizing images in the feature space, fully leveraging feature information and soft labels from the target classifier. Our contributions can be summarized as follows:

\begin{itemize}[leftmargin=0.3cm]
    \item We propose Diffusion-based Model Inversion (Diff-MI) attacks, which capitalize on DMs' remarkable synthesis capabilities to solve MIA problem with high-fidelity reconstruction. To the best of our knowledge, our method is the first diffusion-based white-box MIA.
    \item We devise a target-specific CDM to approximate the target classifier’s private distribution, by creating pseudo-labels as conditions in pretraining and adjusting specified layers with image predictions in fine-tuning. 
    \item We compare existing classification losses and introduce an improved max-margin loss for MIAs consisting of the top-$k$ loss replacing the hard max with top-$k$ maxes and the p-reg loss regularizing images in the feature space.
    \item We conduct extensive experiments to demonstrate that our Diff-MI achieves superior generative fidelity and reconstruction quality with competitive attack accuracy against state-of-the-art (SOTA) methods.
\end{itemize}

\section{Related Works}

\textbf{MIAs} were first proposed in \cite{fredrikson2014privacy,fredrikson2015model}, where they demonstrated success in low-capacity models like logistic regression. Recently, MIAs were extended to tackle more complex DNNs and intended to acquire higher-dimensional private data (e.g., images). The first to launch MIAs against DNNs was Generative Model Inversion (GMI) \cite{zhang2020secret}, where the public dataset was firstly extracted as an image prior by a GAN, and then this prior was employed to guide the reconstruction of private training samples with the guidance of the target classifier. Knowledge-Enriched Distributional Model Inversion (KED-MI) \cite{chen2021knowledge} trained an inversion-specific GAN by introducing soft labels produced by the target classifier and optimizes distributional estimation instead of point estimation in the latent space. Variational Model Inversion (VMI) \cite{wang2021variational} proposed to view the MIA problem as a variational inference problem, and provided a framework using deep normalizing flows \cite{kingma2018glow} regularized by a KL divergence term. Plug \& Play Attacks (PPA) \cite{struppek2022plug} proposed to use publicly available pretrained GANs to attack a wide range of target models. Pseudo Label-Guided Model Inversion (PLG-MI) \cite{yuan2023pseudo} proposed to leverage pseudo-labels produced by the target classifier to guide the training of a conditional GAN \cite{mirza2014conditional} and achieved SOTA attack performance. However, above GAN-based MIAs struggle to achieve a satisfactory balance between attack accuracy and fidelity, either prioritizing high accuracy at the expense of fidelity \cite{zhang2020secret,chen2021knowledge,yuan2023pseudo} or ensuring fidelity at the cost of accuracy \cite{wang2021variational,struppek2022plug}. Based on this limitation, our goal is to achieve a superior accuracy-fidelity balance. More recently, \cite{liu2023unstoppable} introduced conditional diffusion models to tackle black-box MIAs.

\textbf{Diffusion Models} \cite{ho2020denoising} are probabilistic models designed to learn an underlying data distribution and have recently achieved impressive performance in image synthesis tasks, such as conditional generation \cite{dhariwal2021diffusion,ho2022classifier,rombach2022high}, image editing \cite{hertz2022prompt,brooks2023instructpix2pix,li2023zone}, and text-to-image generation \cite{ramesh2021zero,nichol2021glide}. These models employ a Markov chain of diffusion steps, gradually introducing random noise to the data and then learning to reverse this diffusion process to generate desired data samples from noise. To accomplish this, they train a time-conditioned U-Net $\boldsymbol{\epsilon}_{\theta}\left(\mathbf{x}_{t}, t\right)$ \cite{ronneberger2015u} to predict a denoised variant of their input $\mathbf{x}_t$, where $\mathbf{x}_t$ is a noisy version of the input $\mathbf{x}_0$. The main objective can be simplified to 
\begin{equation} \label{eq:1}
\mathcal{L}_{\text{DM}}=\mathbb{E}_{\mathbf{x}_0, \boldsymbol{\epsilon} \sim \mathcal{N}(\mathbf{0}, \mathbf{I}), t}\left[\left\|\boldsymbol{\epsilon}-\boldsymbol{\epsilon}_{\theta}\left(\mathbf{x}_{t}, t\right)\right\|_{2}^{2}\right],
\end{equation}
with $t$ uniformly sampled from $\{1,...,T\}$.

\section{Method}

In this section, we first introduce the MIA problem and then elaborate on the design details of the proposed Diffusion-based Model Inversion (Diff-MI) attacks. 

\subsection{MIA Problem}

\textbf{Problem Formulation.} The goal of MIAs is to construct an attack model, often termed the ``attacker", which is used to reconstruct representative images associated with specified ``target labels" from the ``private dataset" $\mathcal{D}_{\text{pri}}$ with only access to the ``target classifier" $T$. Take face recognition attacks as an example, the attacker can reconstruct a facial image, exposing the privacy of a target label $y_c$ (i.e., person ID) by fully accessing a face recognition classifier. Typically, the reconstructed image should visually and semantically resemble the class-wise training samples of the target classifier.

\textbf{Attacker's Prior Knowledge.} As our method deals with the white-box MIAs, the attacker is granted access to all parameters of the target classifier. The attacker is also presumed to know the data type (e.g., human face) of the private dataset. Therefore, to familiarize the attacker with the private data type, a ``public dataset" $\mathcal{D}_{\text{pub}}$ of the same data type is utilized to gain prior knowledge of general data features, sharing no interclass overlap with $\mathcal{D}_{\text{pri}}$.

\begin{figure*}[t]
    \centering
    \includegraphics[width=1.00\textwidth]{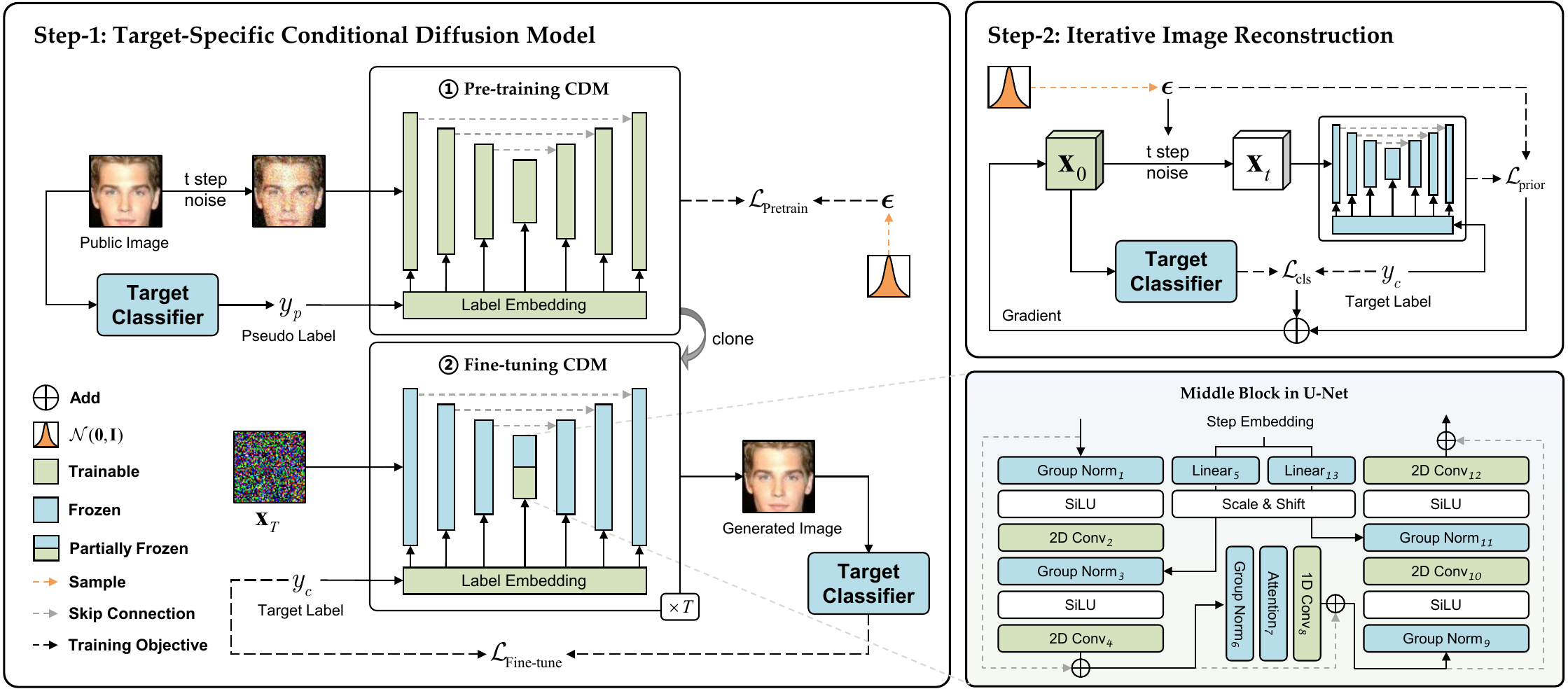}
    \caption{\textbf{Overview of our proposed two-step Diff-MI attacks.} Step-1: We build a target-specific CDM on the public dataset to distill the target classifier's knowledge. This is achieved by pretraining a CDM with pseudo-labels produced by the target classifier as conditions, then fine-tuning a small subset of the pretrained CDM with the guidance of the target classifier (subscripts of layers in the middle block indicate their corresponding index). Step-2: We use an iterative image reconstruction method to involve both diffusion prior (i.e., $\mathcal{L}_{\text{prior}}$) and target classifier's knowledge (i.e., $\mathcal{L}_{\text{cls}}$). $\mathbf{x}_{T}$ in step-1 and $\mathbf{x}_{0}$ in step-2 are both initialized from $\mathcal{N}(\mathbf{0}, \mathbf{I})$.}
    \label{fig:2}
\end{figure*}

\subsection{Diffusion-based Model Inversion Attacks}

An overview of our proposed Diff-MI is illustrated in Fig.~\ref{fig:2}. Given a target classifier $T$ trained on a private dataset $\mathcal{D}_{\text{pri}}$, the attacker aims to reconstruct key features of identities in $\mathcal{D}_{\text{pri}}$ with the support of a public dataset $\mathcal{D}_{\text{pub}}$. Our Diff-MI consists of two steps. In step-1, we build a target-specific CDM to distill knowledge from the target classifier by pretraining a CDM on $\mathcal{D}_{\text{pub}}$ with pseudo-labels as conditions and then fine-tuning a small subset of the pretrained CDM by incorporating the target classifier through its sampling process. In step-2, we propose an iterative image reconstruction method to involve the diffusion prior and the target classifier's knowledge through a combinatorial optimization problem.

\vspace{0.2cm}
\begin{itemize}[leftmargin=0.68cm]
    \item[\textbf{B.1}] \textbf{Step-1: Target-Specific Conditional Diffusion Model} 
\end{itemize}
\vspace{0.2cm}

\textbf{Pre-training CDM.} To leverage the knowledge of the target classifier, we adopt a top-$n$ selection strategy \cite{yuan2023pseudo} in advance to generate pseudo-labels for images in $\mathcal{D}_{\text{pub}}$. Then we utilize the obtained image-label pairs to pretrain the CDM. Specifically, for a provided target classifier $T$ and specified target label $y_c \in \{1, ..., C\}$, where $C$ denotes the number of classes in $\mathcal{D}_{\text{pri}}$, we feed all public images into $T$ and sort the $y_c^{\text{th}}$ value of logits (i.e., outputs of the last layer of $T$) in descending order. Then we select top-$n$ public images with the highest likelihood associated with $y_c$.

This strategy aims to search for the most representative $n$ images in $\mathcal{D}_{\text{pub}}$ corresponding to every target label in $\mathcal{D}_{\text{pri}}$. Empirically, the target classifier has relatively high confidence of class $y_c$ in the selected top-$n$ images, which means these images are perceived to share structural similarity with images of the same class in $\mathcal{D}_{\text{pri}}$ from the perspective of $T$. In this way, the image distribution can be decoupled into distinct class-wise sub-distributions in advance, distilling the knowledge of the target classifier into the CDM. Formally, for training the CDM on $\mathcal{D}_{\text{pub}}$ with pseudo-labels $y_p \in \{1, ..., C\}$, we have a label-conditioned objective via Eq.~\ref{eq:1}, which now reads
\begin{equation}
\mathcal{L}_{\text{Pretrain}}=\mathbb{E}_{\mathbf{x}_0, y_p, \boldsymbol{\epsilon} \sim \mathcal{N}(\mathbf{0}, \mathbf{I}), t}\left[\left\|\boldsymbol{\epsilon}-\boldsymbol{\epsilon}_{\theta}\left(\mathbf{x}_{t}, y_p, t\right)\right\|_{2}^{2}\right].
\end{equation}

\textbf{Fine-tuning CDM.} The use of pseudo-labels is effective in distilling the knowledge of the target classifier. However, we argue that the extracted target knowledge could be inadequate for the attacker, especially when there is a significant gap between the public and private datasets. This inadequacy may arise because the pseudo-labels might not be as accurate as anticipated due to the domain gap. Considering the white-box knowledge of the target classifier (i.e., all parameters are accessible), we propose to integrate the target classifier into the CDM's learning process to further extract its knowledge for our target-specific CDM. In practice, we connect the target classifier to the CDM and initiate a sampling process to generate images. These generated images are then directly fed into the target classifier for label prediction. By comparing the predicted soft labels with the target labels, we can jointly fine-tune the label embedding layer and the U-Net module of the CDM using the following objective
\begin{equation} \label{eq:3}
\mathcal{L}_{\text{Fine-tune}}=\mathbb{E}_{\mathbf{x}_T \sim \mathcal{N}(\mathbf{0}, \mathbf{I}), y_c}\left[\mathcal{L}_{\text{cls}}\left(T\left(D\left(\mathbf{x}_T, y_c\right)\right), y_c\right)\right],
\end{equation}
where $y_c\in \{1, ..., C\}$ represents the target label, $D\left(\cdot, \cdot\right)$ denotes the sampling process of the CDM, and $\mathcal{L}_{\text{cls}}$ is the classification loss we will introduce in Sec.~\ref{sec:loss}.

The most straightforward way for fine-tuning is to adjust all layers of the CDM during the sampling process. However, this operation would undermine the learned mathematical properties of the CDM derived from the diffusion process (i.e., noise-adding process), leading to meaningless generations. Because this contradicts the logic of diffusion models that predict Gaussian noise. Our experiments reveal that only fine-tuning the label embedding layer and the middle block of the U-Net can result in higher attack accuracy but compromise slight generative fidelity. To further minimize the impact on generative fidelity, we adopt the method from \cite{li2020few} to calculate the change rate of layer's parameters using $\Delta_{l}=\left\|\theta_{l}^{\prime}-\theta_{l}\right\| /\left\|\theta_{l}\right\|$, where $\theta_{l}^{\prime}$ and $\theta_{l}$ denote the fine-tuned and original parameters of layer $l$ respectively. As demonstrated by \cite{li2020few}, layers with greater changes typically play a more significant role in fine-tuning. As shown in Fig.~\ref{fig:4} (a), we empirically select five layers in the middle block that show the highest $\Delta$ during fine-tuning, achieving superior accuracy-fidelity balance. Further analysis is provided in Sec.~\ref{sec:ablation_study}.

\vspace{0.2cm}
\begin{itemize}[leftmargin=0.68cm]
    \item[\textbf{B.2}] \textbf{Step-2: Iterative Image Reconstruction} 
\end{itemize}
\vspace{0.2cm}

Upon completing the training of our target-specific CDM, the subsequent step is to employ it for reconstructing images based on a specific target class. Instead of using the traditional diffusion sampling process for image generation, we introduce an iterative image reconstruction (IIR) method, inspired by \cite{graikos2022diffusion}. Particularly, IIR turns the diffusion sampling process into a combinatorial optimization problem, which performs image reconstruction by iterating differentiation involving both the learned CDM prior and target classifier's knowledge. This enables the target classifier to further aid in guiding the image reconstruction process. 

Specifically, given a target label $y_c$, we first initialize the image $\mathbf{x}_0$ from $\mathcal{N}(\mathbf{0}, \mathbf{I})$. Meanwhile, we define a time schedule $(t_i)_{i=1}^{N}$, where $N$ represents the number of iterations, and $t$ is annealed from high to low values. To optimize $\mathbf{x}_0$ under the trained target-specific CDM $\boldsymbol{\epsilon}_{\theta}$ and the constraint of the target classifier $T$, we minimize the following objective consisting of a prior loss $\mathcal{L}_{\text{prior}}$ and a classification loss $\mathcal{L}_{\text{cls}}$:
\begin{equation}
\mathcal{L}_{\text{IIR}} = \mathcal{L}_{\text{prior}} + \mathcal{L}_{\text{cls}}\left(T\left(\mathbf{x}_0\right), y_c\right),
\end{equation}
where
\begin{equation}
\mathcal{L}_{\text{prior}} = \mathbb{E}_{\boldsymbol{\epsilon} \sim \mathcal{N}(\mathbf{0}, \mathbf{I}), t_i}\left[\left\|\boldsymbol{\epsilon}-\boldsymbol{\epsilon}_{\theta}\left(\mathbf{x}_{t_{i}}, y_c, t_{i}\right)\right\|_{2}^{2}\right],
\end{equation}
and $\mathbf{x}_{t_{i}}=\sqrt{\bar{\alpha}_{t_{i}}} \mathbf{x}_0 + \sqrt{1-\bar{\alpha}_{t_{i}}} \boldsymbol{\epsilon}$ represents the diffusion process of DMs according to a predefined schedule $\bar{\alpha}_{t}$.

Intuitively, the prior loss $\mathcal{L}_{\text{prior}}$ measures how well the reconstructed image $\mathbf{x}_0$ aligns with the class-wise diffusion prior. The classification loss $\mathcal{L}_{\text{cls}}$ evaluates the discrepancy between the reconstructed image $\mathbf{x}_0$ and the target label $y_c$ according to the target classifier $T$. After iterating $N$ times, we can obtain the reconstructed image $\mathbf{x}_0$ correlated with the target label $y_c$. Additionally, we adopt the targeted Projected Gradient Descent (PGD) method \cite{madry2017towards,santurkar2019image,ganz2021bigroc} to improve the reconstruction after IIR with the target classifier.

\subsection{An Improved Max-Margin Loss} \label{sec:loss}

Previous works \cite{zhang2020secret,chen2021knowledge,wang2021variational} have commonly adopted the cross-entropy loss for classification while they are usually confronting the vanishing gradient problem. This problem has been proven and addressed by introducing novel losses such as Poincaré loss \cite{struppek2022plug}, max-margin loss \cite{yuan2023pseudo}, and identity loss \cite{nguyen2023re}, among which the max-margin loss achieves SOTA attack performance:
\begin{equation} \label{eq:6}
\mathcal{L}_{\text{MM}}(\mathbf{x}, y_c) = -l_{y_c}(\mathbf{x}) + \max _{j \neq y_c} l_{j}(\mathbf{x}),
\end{equation}
where $l_{y_c}$ denotes the logit w.r.t the target label $y_c$.

We propose a top-$k$ loss as an improvement in the max-margin loss by replacing the hard max with top-$k$ maxes, to make full use of the soft label provided by the target classifier for more discriminative reconstruction. This encourages the discovery of the most representative sample while simultaneously distinguishing it from the other most similar $k$ classes, rather than solely emphasizing the most similar one. Our improved top-$k$ loss is shown as follows:
\begin{equation} \label{eq:7}
\mathcal{L}_{\text{top-k}}(\mathbf{x}, y_c, k) = -l_{y_c}(\mathbf{x}) + \underset{j \neq y_c}{\text{max}_{k}} l_{j}(\mathbf{x}).
\end{equation}

Meanwhile, we find the p-reg loss proposed in \cite{nguyen2023re} useful in regularizing reconstructed images in the feature space (i.e., penultimate layer activations of the classifier) for closer feature distance. They derive the regularizing feature $\mathbf{p}_{\text{reg}}$ by sampling from a feature distribution pre-computed from the whole public dataset using the target classifier. In contrast, we introduce prior knowledge of pseudo-labels to estimate the feature centroid of each target label $\mathbf{p}_{\text{reg}, y_c}$, which now reads
\begin{equation} \label{eq:8}
\mathcal{L}_{\text{p-reg}}(\mathbf{p}_\mathbf{x}, y_c) = \left\|\mathbf{p}_\mathbf{x} - \mathbf{p}_{\text{reg}, y_c} \right\|_{2}^{2},
\end{equation}
where $\mathbf{p}_\mathbf{x}$ refers to penultimate layer activations of the target classifier w.r.t $\mathbf{x}$. To sum up, the classification loss adopted in our Diff-MI can be formulated as 
\begin{equation} \label{eq:9}
\mathcal{L}_{\text{cls}} = \mathcal{L}_{\text{top-k}} + \alpha \mathcal{L}_{\text{p-reg}},
\end{equation}
where $\alpha$ is a regularization coefficient.

\section{Experiments}

In this section, we evaluate our Diff-MI in terms of the performance of reconstructing private images from a specific target classifier. The baselines we compare against are GMI \cite{zhang2020secret}, KED-MI \cite{chen2021knowledge}, and PLG-MI \cite{yuan2023pseudo}, which are all GAN-based methods. We also include more comparisons with VMI \cite{wang2021variational} and PPA \cite{struppek2022plug}, which prioritize fidelity but fall short in accuracy.

\subsection{Experimental Setup}
To ensure a fair comparison, we maintain the same experimental setup for these baselines. Subsequently, we illustrate the details of these setups.

\textbf{Datasets.} We perform MIAs for various classification tasks, including face recognition, fine-grained image classification, and disease prediction. For face recognition, we select three widely-used datasets for experiments: (1) \texttt{CelebA} \cite{liu2015deep} containing 202,599 face images of 10,177 identities with coarse alignment; (2) \texttt{FFHQ} \cite{karras2019style} containing 70,000 high-quality images with considerable variation in terms of age, ethnicity, and image background; and (3) \texttt{FaceScrub} \cite{ng2014data} containing 106,863 face image URLs of 530 celebrities. Herein, we use a cleaned and aligned version of \texttt{FaceScrub} \cite{Khawar_BMVC22_FPVT} which includes 91,712 images of 263 males and 263 females because of partially invalid URLs in the original dataset. For fine-grained image classification, we choose \texttt{CUB-200-2011} \cite{wah2011caltech} which is the most widely-used dataset for fine-grained visual categorization tasks. It contains 11,788 images of 200 subcategories belonging to birds. For disease prediction, \texttt{ChestX-Ray} \cite{wang2017chestx} is a medical image dataset that comprises 112,120 frontal-view X-ray images of 30,805 unique patients with the text-mined fourteen common disease labels, mined from the text radiological reports via NLP techniques. And \texttt{CheXpert} \cite{irvin2019chexpert} contains 224,316 chest radiographs of 65,240 patients with both frontal and lateral views available.

\textbf{Target Classifiers.} Following previous works, we evaluate our attack on three different face recognition models with varying complexities: (1) VGG16 \cite{simonyan2014very}, (2) ResNet-152 (IR152) \cite{he2016deep}, and (3) Face.evoLVe \cite{cheng2017know}. Before training the classifier, the private dataset containing 30,027 images of \texttt{CelebA} is split into two parts: $90\%$ of the samples for training and $10\%$ for testing. All target classifiers are trained using the SGD optimizer with learning rate $10^{-2}$, batch size $64$, momentum $0.9$, and weight decay $10^{-4}$ while making use of pretrained weights. All private images are cropped at the center, resized to $64 \times 64$, and horizontally flipped in $50\%$ of the cases.

\textbf{Implementation Details.} We implement our attack in both standard and distributional shift settings. In the standard setting, we divide the original dataset into two disjoint parts: one serves as $\mathcal{D}_{\text{pri}}$ to train the target classifier and the other serves as $\mathcal{D}_{\text{pub}}$ to train the diffusion model, ensuring there is no image and identity intersection between the two parts. 
% In other words,  $\mathcal{D}_{\text{pub}}$ is only applied to provide auxiliary information (e.g., facial features) for attackers without any overlap of $\mathcal{D}_{\text{pri}}$. 
Specifically, we take 30,027 images of 1,000 identities from \texttt{CelebA} as $\mathcal{D}_{\text{pri}}$ to train the target classifier and randomly choose 30,000 images of other identities from the disjoint part as $\mathcal{D}_{\text{pub}}$ to train the diffusion model. However, the standard setting can be too easy for the current SOTA method \cite{yuan2023pseudo}, thus a more challenging but realistic scenario should be emphasized and researched. In the distributional shift setting, our access is restricted to any part of the original dataset, including its disjoint counterpart. Consequently, we can only resort to another dataset of the same data type, which exhibits a larger distributional shift with $\mathcal{D}_{\text{pri}}$. Herein, we use \texttt{FFHQ} and \texttt{FaceScrub} as substitutes for $\mathcal{D}_{\text{pub}}$. 

\begin{table*}[t]
\caption{\textbf{Attack performance comparison in the standard setting} ($\mathcal{D}_\text{pri}$ = CelebA , $\mathcal{D}_\text{pub}$ = CelebA). ↑ and ↓ respectively symbolize that higher and lower scores give better attack performance.}
\resizebox{\hsize}{!}{
% \tiny  % 调字体大小
\renewcommand{\arraystretch}{1.05}  % 调行距
\begin{tabular}{c|cccc|cccc|cccc}
\toprule
\multirow{2}{*}{\textbf{}} & \multicolumn{4}{c|}{\textbf{VGG16}}            & \multicolumn{4}{c|}{\textbf{IR152}}            & \multicolumn{4}{c}{\textbf{Face.evoLVe}}        \\
               & \textbf{Acc1 ↑}  & \textbf{Acc5 ↑}  & \textbf{FID ↓} & \textbf{KNN Dist ↓} & \textbf{Acc1 ↑}  & \textbf{Acc5 ↑}  & \textbf{FID ↓} & \textbf{KNN Dist ↓} & \textbf{Acc1 ↑}  & \textbf{Acc5 ↑}  & \textbf{FID ↓} & \textbf{KNN Dist ↓} \\ \midrule
GMI         & 23.40\%    & 47.07\%    & 28.04    & 1272.20       & 35.87\%    & 58.53\%    & 29.03    & 1269.30       & 30.87\%    & 53.33\%    & 31.13    & 1297.40       \\
KED-MI    & 63.13\%    & 88.33\%    & 30.49    & 1232.97       & 68.53\%    & 88.07\%    & 41.10    & 1249.90       & 75.00\%    & 94.67\%    & 33.21    & 1232.97       \\
PLG-MI       & \textbf{97.47\%} & \textbf{99.47\%} & 33.27    & 1133.36       & \textbf{99.67\%} & 99.73\% & 33.16    & 1044.64       & \textbf{99.67\%} & \textbf{99.93\%} & 31.48    & 1113.21       \\ \midrule \rowcolor[HTML]{dadada}
\textbf{Ours}                      & 93.47\%    & 99.20\%    & \textbf{23.82} & \textbf{1081.98}    & 97.40\%    & \textbf{99.80\%}    & \textbf{25.77} & \textbf{1010.70}    & 94.93\%    & 99.33\%    & \textbf{28.16} & \textbf{1025.36}    \\ \bottomrule 
\end{tabular}
}
\label{table:1}
\end{table*}

\begin{table*}[t]
\caption{\textbf{Attack performance comparison in the distributional shift setting}. ``$A \to B$"~ represents the diffusion model and target classifier trained on datasets $A$ and $B$, respectively.}
\resizebox{\hsize}{!}{
\tiny  % 调字体大小
\begin{tabular}{cc|cccc|cccc}
\toprule
                                      &                 & \multicolumn{4}{c|}{\textbf{FFHQ → CelebA}}                                 & \multicolumn{4}{c}{\textbf{FaceScrub → CelebA}}                            \\
                                      &                 & \textbf{Acc1 ↑}  & \textbf{Acc5 ↑}  & \textbf{FID ↓} & \textbf{KNN Dist ↓} & \textbf{Acc1 ↑}  & \textbf{Acc5 ↑}  & \textbf{FID ↓} & \textbf{KNN Dist ↓} \\ \midrule
\multirow{4}{*}{\textbf{VGG16}}       & GMI    & 8.40\%           & 21.07\%          & 41.55          & 1406.68             & 11.07\%          & 27.87\%          & 39.47          & 1386.05             \\
                                      & KED-MI & 33.93\%          & 64.93\%          & 37.37          & 1353.46             & 39.13\%          & 70.33\%          & 36.90          & 1344.16             \\
                                      & PLG-MI & \textbf{87.07\%} & \textbf{95.73\%} & 43.55          & 1277.54             & \textbf{83.00\%} & \textbf{95.93\%} & 34.08          & 1247.40             \\ \rowcolor[HTML]{dadada}
\cellcolor{white}                     & \textbf{Ours}   & 78.07\%          & 93.87\%          & \textbf{28.82} & \textbf{1250.04}    & 75.13\%          & 92.40\%          & \textbf{25.74} & \textbf{1221.25}    \\ \midrule
\multirow{4}{*}{\textbf{IR152}}       & GMI    & 14.93\%          & 33.13\%          & 41.87          & 1402.52             & 18.93\%          & 41.00\%          & 41.98          & 1340.84             \\
                                      & KED-MI & 44.60\%          & 74.07\%          & 46.97          & 1320.22             & 52.27\%          & 78.80\%          & 33.97          & 1307.37             \\
                                      & PLG-MI & \textbf{96.67\%} & \textbf{99.67\%} & 44.15          & 1159.10             & \textbf{96.73\%} & \textbf{99.33\%} & 40.34          & 1137.90             \\ \rowcolor[HTML]{dadada}
\cellcolor{white}                     & \textbf{Ours}   & 94.73\%          & \textbf{99.67\%}          & \textbf{37.82} & \textbf{1140.09}    & 90.40\%          & 97.67\%          & \textbf{29.53} & \textbf{1113.21}    \\ \midrule
\multirow{4}{*}{\textbf{Face.evoLVe}} & GMI    & 11.33\%          & 26.33\%          & 46.29          & 1393.91             & 14.00\%          & 31.67\%          & 42.05          & 1371.37             \\
                                      & KED-MI & 48.47\%          & 74.07\%          & 44.53          & 1336.24             & 46.07\%          & 75.27\%          & 39.75          & 1335.63             \\
                                      & PLG-MI & \textbf{93.13\%} & \textbf{98.60\%} & 48.45          & 1231.42             & \textbf{91.87\%} & \textbf{98.33\%} & 40.00          & 1245.22             \\ \rowcolor[HTML]{dadada}
\cellcolor{white}                     & \textbf{Ours}   & 92.60\%          & \textbf{98.60\%}          & \textbf{37.73} & \textbf{1204.60}    & 89.53\%          & 98.00\%          & \textbf{27.44} & \textbf{1210.82}    \\ \bottomrule
\end{tabular}
}
\label{table:2}
\end{table*}

\begin{table*}[t]
\caption{\textbf{Reconstruction quality assessment using different methods with various target classifiers.} LPIPS-A and LPIPS-V denote that using AlexNet \protect\cite{krizhevsky2012imagenet} and VGG16 \protect\cite{simonyan2014very} for extracting image features, respectively. ``$A + B$" indicates taking $A$ as the public dataset and $B$ as the target classifier.}
\resizebox{\hsize}{!}{
% \scriptsize  % 调字体大小
\renewcommand{\arraystretch}{1.20}  % 调行距
\begin{tabular}{ccccccccccccc}
\toprule
                            &                                                                                                         &                 & \textbf{PSNR ↑} & \textbf{SSIM ↑} & \textbf{FSIM ↑} & \textbf{VSI ↑} & \textbf{HaarPSI ↑} & \textbf{SR-SIM ↑} & \textbf{DSS ↑} & \textbf{MDSI ↓} & \textbf{LPIPS-A ↓} & \textbf{LPIPS-V ↓} \\ \midrule
\multirow{4}{*}{\textbf{1}} & \multirow{4}{*}{\textbf{\begin{tabular}[c]{@{}c@{}}CelebA\\      +\\      VGG16\end{tabular}}}          & GMI    & 13.09           & 0.39            & 0.69            & 0.86           & 0.40               & 0.82              & 0.44           & 0.57            & 0.3393             & 0.5245             \\
                            &                                                                                                         & KED-MI & 14.13           & 0.42            & 0.71            & 0.86           & 0.42               & 0.84              & 0.51           & 0.56            & 0.3324             & 0.5175             \\
                            &                                                                                                         & PLG-MI & 14.79           & 0.47            & 0.73            & 0.88           & 0.45               & 0.85              & 0.52           & 0.54            & 0.3025             & 0.4787             \\
                            &                                                                                                         & \cellcolor[HTML]{dadada}\textbf{Ours}   & \cellcolor[HTML]{dadada}\textbf{15.64}  & \cellcolor[HTML]{dadada}\textbf{0.53}   & \cellcolor[HTML]{dadada}\textbf{0.75}   & \cellcolor[HTML]{dadada}\textbf{0.89}  & \cellcolor[HTML]{dadada}\textbf{0.49}      & \cellcolor[HTML]{dadada}\textbf{0.87}     & \cellcolor[HTML]{dadada}\textbf{0.60}  & \cellcolor[HTML]{dadada}\textbf{0.53}   & \cellcolor[HTML]{dadada}\textbf{0.2998}    & \cellcolor[HTML]{dadada}\textbf{0.4747}    \\ \midrule
\multirow{4}{*}{\textbf{2}} & \multirow{4}{*}{\textbf{\begin{tabular}[c]{@{}c@{}}CelebA\\      +\\      IR152\end{tabular}}}          & GMI    & 13.15           & 0.38            & 0.69            & 0.86           & 0.40               & 0.82              & 0.44           & 0.56            & 0.3406             & 0.5259             \\
                            &                                                                                                         & KED-MI & 13.94           & 0.38            & 0.69            & 0.86           & 0.41               & 0.83              & 0.51           & 0.56            & 0.3414             & 0.5322             \\
                            &                                                                                                         & PLG-MI & 14.62           & 0.49            & 0.73            & 0.88           & 0.44               & 0.84              & 0.53           & 0.55            & 0.3085             & 0.4855             \\
                            &                                                                                                         & \cellcolor[HTML]{dadada}\textbf{Ours}   & \cellcolor[HTML]{dadada}\textbf{16.11}  & \cellcolor[HTML]{dadada}\textbf{0.55}   & \cellcolor[HTML]{dadada}\textbf{0.76}   & \cellcolor[HTML]{dadada}\textbf{0.90}  & \cellcolor[HTML]{dadada}\textbf{0.50}      & \cellcolor[HTML]{dadada}\textbf{0.86}     & \cellcolor[HTML]{dadada}\textbf{0.61}  & \cellcolor[HTML]{dadada}\textbf{0.54}   & \cellcolor[HTML]{dadada}\textbf{0.2991}    & \cellcolor[HTML]{dadada}\textbf{0.4791}    \\ \midrule
\multirow{4}{*}{\textbf{3}} & \multirow{4}{*}{\textbf{\begin{tabular}[c]{@{}c@{}}CelebA\\      +\\      Face.evoLVe\end{tabular}}}    & GMI    & 12.93           & 0.38            & 0.69            & 0.86           & 0.39               & 0.82              & 0.43           & 0.57            & 0.3434             & 0.5306             \\
                            &                                                                                                         & KED-MI & 13.95           & 0.42            & 0.70            & 0.86           & 0.41               & 0.84              & 0.52           & 0.56            & 0.3340             & 0.5200             \\
                            &                                                                                                         & PLG-MI & 14.20           & 0.48            & 0.72            & 0.87           & 0.43               & 0.84              & 0.53           & 0.55            & 0.3069             & 0.4859             \\
                            &                                                                                                         & \cellcolor[HTML]{dadada}\textbf{Ours}   & \cellcolor[HTML]{dadada}\textbf{16.31}  & \cellcolor[HTML]{dadada}\textbf{0.55}   & \cellcolor[HTML]{dadada}\textbf{0.76}   & \cellcolor[HTML]{dadada}\textbf{0.90}  & \cellcolor[HTML]{dadada}\textbf{0.50}      & \cellcolor[HTML]{dadada}\textbf{0.87}     & \cellcolor[HTML]{dadada}\textbf{0.60}  & \cellcolor[HTML]{dadada}\textbf{0.54}   & \cellcolor[HTML]{dadada}\textbf{0.2975}    & \cellcolor[HTML]{dadada}\textbf{0.4737}    \\ \midrule
\multirow{4}{*}{\textbf{4}} & \multirow{4}{*}{\textbf{\begin{tabular}[c]{@{}c@{}}FFHQ\\      +\\      VGG16\end{tabular}}}            & GMI    & 11.86           & 0.22            & 0.65            & 0.83           & 0.35               & 0.80              & 0.36           & 0.58            & 0.3442             & 0.5502             \\
                            &                                                                                                         & KED-MI & 11.31           & 0.24            & 0.64            & 0.81           & 0.34               & 0.79              & 0.38           & 0.58            & 0.3622             & 0.5532             \\
                            &                                                                                                         & PLG-MI & 13.49           & 0.28            & 0.69            & 0.85           & 0.40               & 0.82              & 0.44           & 0.55            & 0.3108             & 0.5092             \\
                            &                                                                                                         & \cellcolor[HTML]{dadada}\textbf{Ours}   & \cellcolor[HTML]{dadada}\textbf{14.07}  & \cellcolor[HTML]{dadada}\textbf{0.33}   & \cellcolor[HTML]{dadada}\textbf{0.70}   & \cellcolor[HTML]{dadada}\textbf{0.86}  & \cellcolor[HTML]{dadada}\textbf{0.43}      & \cellcolor[HTML]{dadada}\textbf{0.84}     & \cellcolor[HTML]{dadada}\textbf{0.48}  & \cellcolor[HTML]{dadada}\textbf{0.54}   & \cellcolor[HTML]{dadada}\textbf{0.3069}    & \cellcolor[HTML]{dadada}\textbf{0.5024}    \\ \midrule
\multirow{4}{*}{\textbf{5}} & \multirow{4}{*}{\textbf{\begin{tabular}[c]{@{}c@{}}FFHQ\\      +\\      IR152\end{tabular}}}            & GMI    & 11.77           & 0.23            & 0.65            & 0.83           & 0.35               & 0.79              & 0.36           & 0.58            & 0.3490             & 0.5544             \\
                            &                                                                                                         & KED-MI & 11.00           & 0.23            & 0.63            & 0.81           & 0.33               & 0.79              & 0.36           & 0.58            & 0.3685             & 0.5602             \\
                            &                                                                                                         & PLG-MI & 13.69           & 0.30            & \textbf{0.70}            & \textbf{0.86}           & 0.40               & 0.82              & 0.43           & 0.55            & 0.3178             & 0.5115             \\
                            &                                                                                                         & \cellcolor[HTML]{dadada}\textbf{Ours}   & \cellcolor[HTML]{dadada}\textbf{13.72}           & \cellcolor[HTML]{dadada}\textbf{0.31}            & \cellcolor[HTML]{dadada}0.69            & \cellcolor[HTML]{dadada}0.85           & \cellcolor[HTML]{dadada}\textbf{0.41}               & \cellcolor[HTML]{dadada}\textbf{0.83}              & \cellcolor[HTML]{dadada}\textbf{0.46}           & \cellcolor[HTML]{dadada}\textbf{0.54}            & \cellcolor[HTML]{dadada}\textbf{0.3124}             & \cellcolor[HTML]{dadada}\textbf{0.5103}             \\ \midrule
\multirow{4}{*}{\textbf{6}} & \multirow{4}{*}{\textbf{\begin{tabular}[c]{@{}c@{}}FFHQ\\      +\\      Face.evoLVe\end{tabular}}}      & GMI    & 11.62           & 0.22            & 0.65            & 0.83           & 0.34               & 0.79              & 0.35           & 0.58            & 0.3506             & 0.5595             \\
                            &                                                                                                         & KED-MI & 11.06           & 0.24            & 0.63            & 0.81           & 0.34               & 0.79              & 0.37           & 0.58            & 0.3576             & 0.5466             \\
                            &                                                                                                         & PLG-MI & 13.04           & 0.28            & 0.68            & 0.85           & 0.38               & 0.82              & 0.42           & 0.56            & 0.3255             & 0.5169             \\
                            &                                                                                                         & \cellcolor[HTML]{dadada}\textbf{Ours}   & \cellcolor[HTML]{dadada}\textbf{13.80}  & \cellcolor[HTML]{dadada}\textbf{0.33}   & \cellcolor[HTML]{dadada}\textbf{0.70}   & \cellcolor[HTML]{dadada}\textbf{0.86}  & \cellcolor[HTML]{dadada}\textbf{0.43}      & \cellcolor[HTML]{dadada}\textbf{0.83}     & \cellcolor[HTML]{dadada}\textbf{0.48}  & \cellcolor[HTML]{dadada}\textbf{0.54}   & \cellcolor[HTML]{dadada}\textbf{0.3116}    & \cellcolor[HTML]{dadada}\textbf{0.5037}    \\ \midrule
\multirow{4}{*}{\textbf{7}} & \multirow{4}{*}{\textbf{\begin{tabular}[c]{@{}c@{}}FaceScrub\\      +\\      VGG16\end{tabular}}}       & GMI    & 12.97           & 0.28            & 0.68            & \textbf{0.85}  & 0.38               & 0.82              & 0.41           & 0.56            & 0.3402             & 0.5426             \\
                            &                                                                                                         & KED-MI & 11.74           & 0.26            & 0.65            & 0.82           & 0.36               & 0.80              & 0.40           & 0.57            & 0.3498             & 0.5445             \\
                            &                                                                                                         & PLG-MI & \textbf{13.67}  & 0.29            & 0.69            & \textbf{0.85}  & \textbf{0.41}      & 0.82              & 0.44           & 0.55            & 0.3147             & \textbf{0.5090}    \\
                            &                                                                                                         & \cellcolor[HTML]{dadada}\textbf{Ours}   & \cellcolor[HTML]{dadada}13.63           & \cellcolor[HTML]{dadada}\textbf{0.33}   & \cellcolor[HTML]{dadada}\textbf{0.70}   & \cellcolor[HTML]{dadada}\textbf{0.85}  & \cellcolor[HTML]{dadada}\textbf{0.41}      & \cellcolor[HTML]{dadada}\textbf{0.83}     & \cellcolor[HTML]{dadada}\textbf{0.47}  & \cellcolor[HTML]{dadada}\textbf{0.54}   & \cellcolor[HTML]{dadada}\textbf{0.3078}    & \cellcolor[HTML]{dadada}0.5129             \\ \midrule
\multirow{4}{*}{\textbf{8}} & \multirow{4}{*}{\textbf{\begin{tabular}[c]{@{}c@{}}FaceScrub\\      +\\      IR152\end{tabular}}}       & GMI    & 12.89           & 0.28            & 0.68            & 0.85           & 0.38               & 0.82              & 0.40           & 0.57            & 0.3436             & 0.5458             \\
                            &                                                                                                         & KED-MI & 11.93           & 0.26            & 0.65            & 0.83           & 0.36               & 0.81              & 0.40           & 0.57            & 0.3504             & 0.5463             \\
                            &                                                                                                         & PLG-MI & 13.42           & 0.30            & 0.69            & 0.85           & 0.41               & 0.82              & 0.44           & 0.55            & 0.3165             & \textbf{0.5111}    \\
                            &                                                                                                         & \cellcolor[HTML]{dadada}\textbf{Ours}   & \cellcolor[HTML]{dadada}\textbf{13.82}  & \cellcolor[HTML]{dadada}\textbf{0.34}   & \cellcolor[HTML]{dadada}\textbf{0.70}   & \cellcolor[HTML]{dadada}\textbf{0.86}  & \cellcolor[HTML]{dadada}\textbf{0.42}      & \cellcolor[HTML]{dadada}\textbf{0.83}     & \cellcolor[HTML]{dadada}\textbf{0.48}  & \cellcolor[HTML]{dadada}\textbf{0.54}   & \cellcolor[HTML]{dadada}\textbf{0.3060}    & \cellcolor[HTML]{dadada}0.5130             \\ \midrule
\multirow{4}{*}{\textbf{9}} & \multirow{4}{*}{\textbf{\begin{tabular}[c]{@{}c@{}}FaceScrub\\      +\\      Face.evoLVe\end{tabular}}} & GMI    & 12.74           & 0.27            & 0.67            & \textbf{0.85}  & 0.37               & 0.82              & 0.40           & 0.56            & 0.3479             & 0.5498             \\
                            &                                                                                                         & KED-MI & 11.70           & 0.25            & 0.65            & 0.82           & 0.35               & 0.80              & 0.39           & 0.57            & 0.3484             & 0.5427             \\
                            &                                                                                                         & PLG-MI & 13.46           & 0.30            & 0.68            & \textbf{0.85}  & 0.40               & 0.82              & 0.43           & 0.55            & 0.3153             & 0.5177             \\
                            &                                                                                                         & \cellcolor[HTML]{dadada}\textbf{Ours}   & \cellcolor[HTML]{dadada}\textbf{13.70}  & \cellcolor[HTML]{dadada}\textbf{0.33}   & \cellcolor[HTML]{dadada}\textbf{0.69}   & \cellcolor[HTML]{dadada}\textbf{0.85}  & \cellcolor[HTML]{dadada}\textbf{0.42}      & \cellcolor[HTML]{dadada}\textbf{0.83}     & \cellcolor[HTML]{dadada}\textbf{0.47}  & \cellcolor[HTML]{dadada}\textbf{0.54}   & \cellcolor[HTML]{dadada}\textbf{0.3127}    & \cellcolor[HTML]{dadada}\textbf{0.5100}    \\ \bottomrule
\end{tabular}
}
\label{table:resconstruction_assessment}
\end{table*}

\subsection{Evaluation Metrics} \label{sec:discussion_on_metrics_appendix}
The evaluation of MIAs can be multifaceted, with the core aim of assessing whether the reconstructed images expose any private information about the target class. We reconstruct 5 images for the first 300 target labels for evaluation. Moreover, we conduct a user study to access generative fidelity and accuracy from the perspective of human preference. Herein, we introduce metrics to meet the needs of visual inspection as well as quantitative evaluation: 

\textbf{(1) Attack Accuracy (Acc).}  We build an \textit{evaluation classifier} to predict target classes of the reconstruction and then calculate top-$1$ (Acc1) and top-$5$ (Acc5) accuracies. Notably, the evaluation classifier should be different from the target classifier, as reconstructed images may incorporate visually irrelevant and semantically meaningless features, potentially leading to overfitting and inflated attack accuracy. Moreover, the evaluation classifier should achieve high generalization ability because it serves as a proxy for a human observer in determining whether the reconstruction reveals personally sensitive information. Herein, we adopt the model in \cite{cheng2017know}, which is pretrained on \texttt{MS-Celeb-1M} \cite{guo2016ms} and fine-tuned on the same training data of the target classifier.

\textbf{(2) Fréchet Inception Distance (FID).} FID \cite{heusel2017gans} is commonly used for evaluating the generative diversity and fidelity in GAN's works. It measures feature distances between generated images and real images. Feature vectors are all extracted by an Inception-v3 model \cite{szegedy2016rethinking} pretrained on ImageNet \cite{deng2009imagenet}. In our experiment, we calculate FID between the reconstructed images and the private training set of the target classifier to reveal the generative fidelity of the reconstruction. Lower FID typically indicates that the reconstructed images are resembling the private images with superior generative fidelity. In our implementation, we use Pytorch-FID \cite{Seitzer2020FID}.
% Previous works \cite{chen2021knowledge, yuan2023pseudo} only compute FID on the reconstructed images recognized successfully by the evaluation classifier. In contrast with them, we calculate FID on all reconstructed images as we consider that FID is a statistical metric where all generated images should be taken into consideration to estimate an accurate distribution of the reconstruction. 

\textbf{(3) K-Nearest Neighbor Distance (KNN Dist).} KNN Dist means the shortest feature distance ($\ell_2$ distance) between the reconstructed image for a specific target label and corresponding private images of the same class. Lower KNN Dist typically indicates that the reconstructed images are closer to the private images. Notably, KNN Dist is a model-specific metric correlated with Acc and we found that previous works \cite{zhang2020secret,chen2021knowledge} with relatively lower Acc usually suffer from a much higher KNN Dist, resulting in a misleading evaluation. To achieve independence from Acc and precise evaluations, we propose to only calculate KNN Dist on the reconstructed images recognized successfully by the evaluation classifier to have a fair comparison.

\textbf{(4) PSNR, SSIM, \& LPIPS} \cite{hore2010image,wang2004image,zhang2018unreasonable} are three widely used metrics in the field of image quality assessment. PSNR measures the reconstruction accuracy of an image by comparing the peak signal-to-noise ratio between the reconstructed and private images, with higher values indicating better image quality. SSIM focuses on structural similarity, considering aspects such as brightness, contrast, and structure, with values ranging from -1 to 1, where closer to 1 indicates higher image quality. LPIPS is a deep learning-based metric that assesses image differences by learning the perceptual similarity of image patches, providing a more perceptually aligned evaluation, which has been demonstrated to closely align with human perception. We also introduce additional metrics regarding image quality assessment, i.e., FSIM \cite{zhang2011fsim}, VSI \cite{zhang2014vsi}, HaarPSI \cite{reisenhofer2018haar}, SR-SIM \cite{zhang2012sr}, DSS \cite{balanov2015image}, and MDSI \cite{nafchi2016mean}.

\textbf{Discussion on MIA Metrics.} Notably, we would like to distinguish between the attack accuracy (Acc) and other metrics regarding reconstruction similarity (i.e., FID, KNN Dist, and PSNR/SSIM/LPIPS). Acc relies on an evaluation classifier and is inevitably subject to inherent prediction errors, which indicates that Acc may not always accurately and objectively reflect the attack performance and largely depends on the similarity between the evaluation and target classifiers (e.g., model architecture, data augmentation, and training strategy). In contrast, other metrics directly measure the feature distance or image similarity to private images, resulting in fewer cumulative errors and a more objective assessment. Our user study also shows a preference for higher-fidelity images, even if they achieve lower Acc. This implies that once Acc is sufficiently high (e.g., when the user's key privacy is successfully reconstructed), the image quality becomes more crucial. This perspective prioritizes image quality based on human observation over a purely numerical fit to the classifier. This holistic consideration underscores the necessity of compromising attack accuracy and generative fidelity, i.e., \textbf{superior accuracy-fidelity balance} in our setup.

\subsection{Comparisons on Face Datasets} \label{sec:reconstruction_quality}

\begin{figure}[t]
    \centering
    \includegraphics[width=\hsize]{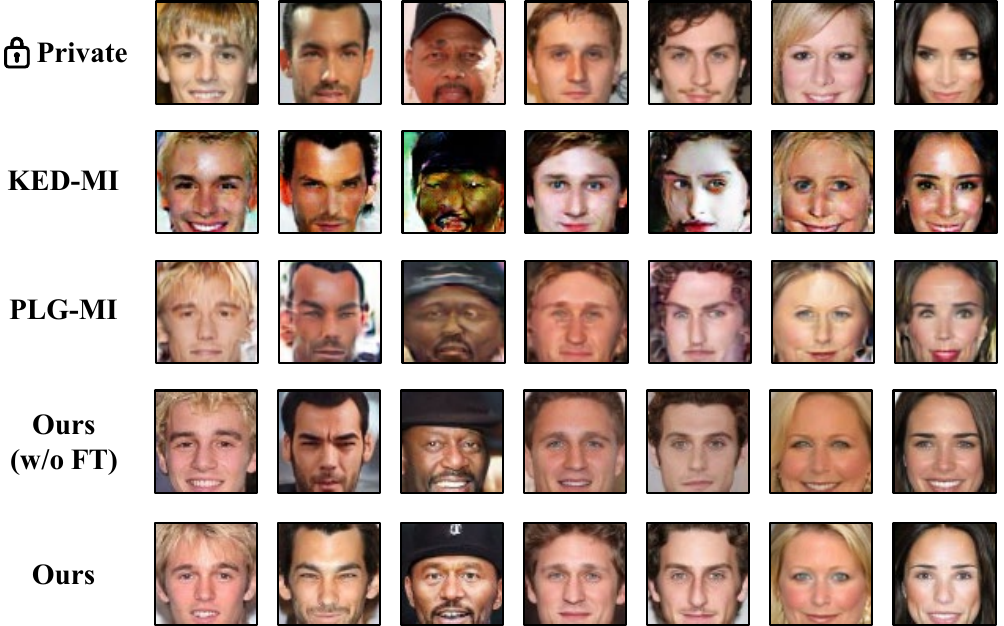}
    \caption{\textbf{Visual comparison of reconstructed images} using different MIA methods ($\mathcal{D}_{\text{pri}}$ = CelebA, $\mathcal{D}_{\text{pub}}$ = CelebA, Target Classifier = VGG16). The first row shows the ground-truth private images of target labels of different identities.}
    \label{fig:3}
\end{figure}

\begin{table*}[t]
\caption{\textbf{Attack performance between Diff-MI and PPA} ($\mathcal{D}_{\text{pri}}$ = CelebA, $\mathcal{D}_{\text{pub}}$ = FFHQ, image size = $64 \times 64$).}
\resizebox{\hsize}{!}{
% \scriptsize  % 调字体大小
\renewcommand{\arraystretch}{1.10}  % 调行距
\begin{tabular}{cccccccccccccccc}
\toprule
                                      &               & \textbf{Acc1 ↑}  & \textbf{Acc5 ↑}  & \textbf{FID ↓} & \textbf{KNN Dist ↓} & \textbf{PSNR ↑} & \textbf{SSIM ↑} & \textbf{FSIM ↑} & \textbf{VSI ↑} & \textbf{HaarPSI ↑} & \textbf{SR-SIM ↑} & \textbf{DSS ↑} & \textbf{MDSI ↓} & \textbf{LPIPS-A ↓} & \textbf{LPIPS-V ↓} \\ \midrule
\multirow{2}{*}{\textbf{VGG16}}       & PPA           & 19.53\%          & 38.13\%          & \textbf{24.79} & \textbf{1230.81}    & 13.84           & 0.28            & 0.69            & \textbf{0.86}  & 0.39               & 0.82              & 0.40           & 0.57            & 0.3460             & 0.5240             \\
                                      & \cellcolor[HTML]{dadada}\textbf{Ours} & \cellcolor[HTML]{dadada}\textbf{78.07\%} & \cellcolor[HTML]{dadada}\textbf{93.87\%} & \cellcolor[HTML]{dadada}28.82          & \cellcolor[HTML]{dadada}1250.04             & \cellcolor[HTML]{dadada}\textbf{14.07}  & \cellcolor[HTML]{dadada}\textbf{0.33}   & \cellcolor[HTML]{dadada}\textbf{0.70}   & \cellcolor[HTML]{dadada}\textbf{0.86}  & \cellcolor[HTML]{dadada}\textbf{0.43}      & \cellcolor[HTML]{dadada}\textbf{0.84}     & \cellcolor[HTML]{dadada}\textbf{0.48}  & \cellcolor[HTML]{dadada}\textbf{0.54}   & \cellcolor[HTML]{dadada}\textbf{0.3069}    & \cellcolor[HTML]{dadada}\textbf{0.5024}    \\ \midrule
\multirow{2}{*}{\textbf{IR152}}       & PPA           & 22.93\%          & 44.27\%          & 24.82          & 1239.97             & 13.70           & 0.27            & \textbf{0.69}   & \textbf{0.86}  & 0.38               & 0.82              & 0.39           & 0.57            & 0.3532             & 0.5290             \\
                                      & \cellcolor[HTML]{dadada}\textbf{Ours} & \cellcolor[HTML]{dadada}\textbf{94.73\%} & \cellcolor[HTML]{dadada}\textbf{99.67\%} & \cellcolor[HTML]{dadada}\textbf{37.82} & \cellcolor[HTML]{dadada}\textbf{1140.09}    & \cellcolor[HTML]{dadada}\textbf{13.72}  & \cellcolor[HTML]{dadada}\textbf{0.31}   & \cellcolor[HTML]{dadada}\textbf{0.69}   & \cellcolor[HTML]{dadada}0.85           & \cellcolor[HTML]{dadada}\textbf{0.41}      & \cellcolor[HTML]{dadada}\textbf{0.83}     & \cellcolor[HTML]{dadada}\textbf{0.46}  & \cellcolor[HTML]{dadada}\textbf{0.54}   & \cellcolor[HTML]{dadada}\textbf{0.3124}    & \cellcolor[HTML]{dadada}\textbf{0.5103}    \\ \midrule
\multirow{2}{*}{\textbf{Face.evoLVe}} & PPA           & 21.87\%          & 43.93\%          & \textbf{27.17} & 1272.65             & 13.68           & 0.27            & 0.69            & \textbf{0.86}  & 0.38               & 0.82              & 0.39           & 0.57            & 0.3512             & 0.5311             \\
                                      & \cellcolor[HTML]{dadada}\textbf{Ours} & \cellcolor[HTML]{dadada}\textbf{92.60\%} & \cellcolor[HTML]{dadada}\textbf{98.60\%} & \cellcolor[HTML]{dadada}37.73          & \cellcolor[HTML]{dadada}\textbf{1204.60}    & \cellcolor[HTML]{dadada}\textbf{13.80}  & \cellcolor[HTML]{dadada}\textbf{0.33}   & \cellcolor[HTML]{dadada}\textbf{0.70}   & \cellcolor[HTML]{dadada}\textbf{0.86}  & \cellcolor[HTML]{dadada}\textbf{0.43}      & \cellcolor[HTML]{dadada}\textbf{0.83}     & \cellcolor[HTML]{dadada}\textbf{0.48}  & \cellcolor[HTML]{dadada}\textbf{0.54}   & \cellcolor[HTML]{dadada}\textbf{0.3116}    & \cellcolor[HTML]{dadada}\textbf{0.5037}    \\ \bottomrule
\end{tabular}
}
\label{table:PPA_appendix}
\end{table*}

\begin{table*}[t]
\caption{\textbf{Attack performance between Diff-MI and VMI} ($\mathcal{D}_{\text{pri}}$ = CelebA, $\mathcal{D}_{\text{pub}}$ = CelebA).}
\resizebox{\hsize}{!}{
% \scriptsize  % 调字体大小
\renewcommand{\arraystretch}{1.10}  % 调行距
\begin{tabular}{cccccccccccccccc}
\toprule
                                      &               & \textbf{Acc1 ↑}  & \textbf{Acc5 ↑}  & \textbf{FID ↓} & \textbf{KNN Dist ↓} & \textbf{PSNR ↑} & \textbf{SSIM ↑} & \textbf{FSIM ↑} & \textbf{VSI ↑} & \textbf{HaarPSI ↑} & \textbf{SR-SIM ↑} & \textbf{DSS ↑} & \textbf{MDSI ↓} & \textbf{LPIPS-A ↓} & \textbf{LPIPS-V ↓} \\ \midrule
\multirow{2}{*}{\textbf{VGG16}}       & VMI           & 54.60\%          & 83.67\%          & \textbf{20.61} & 1272.29    & 14.28           & 0.39            & 0.71            & 0.87  & 0.42               & 0.83              & 0.49           & 0.56            & 0.3133             & 0.4988             \\
                                      & \cellcolor[HTML]{dadada}\textbf{Ours} & \cellcolor[HTML]{dadada}\textbf{93.47\%} & \cellcolor[HTML]{dadada}\textbf{99.20\%} & \cellcolor[HTML]{dadada}23.82          & \cellcolor[HTML]{dadada}\textbf{1081.98}    & \cellcolor[HTML]{dadada}\textbf{15.64}  & \cellcolor[HTML]{dadada}\textbf{0.53}   & \cellcolor[HTML]{dadada}\textbf{0.75}   & \cellcolor[HTML]{dadada}\textbf{0.89}  & \cellcolor[HTML]{dadada}\textbf{0.49}      & \cellcolor[HTML]{dadada}\textbf{0.87}     & \cellcolor[HTML]{dadada}\textbf{0.60}  & \cellcolor[HTML]{dadada}\textbf{0.53}   & \cellcolor[HTML]{dadada}\textbf{0.2998}    & \cellcolor[HTML]{dadada}\textbf{0.4747}    \\ \midrule
\multirow{2}{*}{\textbf{IR152}}       & VMI           & 74.33\%          & 93.33\%          & \textbf{21.50} & 1235.52             & 14.48           & 0.39            & 0.71   & 0.87  & 0.42               & 0.84              & 0.51           & 0.56            & 0.3204             & 0.5071             \\
                                      & \cellcolor[HTML]{dadada}\textbf{Ours} & \cellcolor[HTML]{dadada}\textbf{97.40\%} & \cellcolor[HTML]{dadada}\textbf{99.80\%} & \cellcolor[HTML]{dadada}25.77 & \cellcolor[HTML]{dadada}\textbf{1010.70}    & \cellcolor[HTML]{dadada}\textbf{16.11}  & \cellcolor[HTML]{dadada}\textbf{0.55}   & \cellcolor[HTML]{dadada}\textbf{0.76}   & \cellcolor[HTML]{dadada}\textbf{0.90}  & \cellcolor[HTML]{dadada}\textbf{0.50}      & \cellcolor[HTML]{dadada}\textbf{0.86}     & \cellcolor[HTML]{dadada}\textbf{0.61}  & \cellcolor[HTML]{dadada}\textbf{0.54}   & \cellcolor[HTML]{dadada}\textbf{0.2991}    & \cellcolor[HTML]{dadada}\textbf{0.4791}    \\ \midrule
\multirow{2}{*}{\textbf{Face.evoLVe}} & VMI           & 68.27\% & 91.33\% & \textbf{19.45} & 1231.84    & 14.28  & 0.41   & 0.71   & 0.87  & 0.42      & 0.83     & 0.50  & 0.56   & 0.3138     & 0.4970    \\
                                      & \cellcolor[HTML]{dadada}\textbf{Ours} & \cellcolor[HTML]{dadada}\textbf{94.93\%} & \cellcolor[HTML]{dadada}\textbf{99.33\%} & \cellcolor[HTML]{dadada}28.16          & \cellcolor[HTML]{dadada}\textbf{1025.36}    & \cellcolor[HTML]{dadada}\textbf{16.31}  & \cellcolor[HTML]{dadada}\textbf{0.55}   & \cellcolor[HTML]{dadada}\textbf{0.76}   & \cellcolor[HTML]{dadada}\textbf{0.90}  & \cellcolor[HTML]{dadada}\textbf{0.50}      & \cellcolor[HTML]{dadada}\textbf{0.87}     & \cellcolor[HTML]{dadada}\textbf{0.60}  & \cellcolor[HTML]{dadada}\textbf{0.54}   & \cellcolor[HTML]{dadada}\textbf{0.2975}    & \cellcolor[HTML]{dadada}\textbf{0.4737}    \\ \bottomrule
\end{tabular}
}
\label{table:VMI_appendix}
\end{table*}

\begin{table}[t]
\centering
\vspace{0.2cm}
\caption{\textbf{MIAs in high-resolution} compared with PPA (FFHQ $\to$ CelebA, $T$ = IR152, image size = $224 \times 224$).}
% \resizebox{\hsize}{!}{
\footnotesize  % 调字体大小
\renewcommand{\arraystretch}{1.1}  % 调行距
\setlength\tabcolsep{8pt}  % 调列距
\begin{tabular}{c|cccc}
\toprule
                                      & \textbf{Acc1 ↑}  & \textbf{Acc5 ↑}  & \textbf{FID ↓}     & \textbf{KNN Dist ↓}     \\ \midrule
PPA                                   & 77.26\%          & 93.82\%          & 64.11              & 135.85                  \\ \rowcolor[HTML]{dadada}
\textbf{Ours}                         & \textbf{91.38\%} & \textbf{98.96\%} & \textbf{57.51}     & \textbf{133.48}         \\ \bottomrule
\end{tabular}
% }
\label{table:high_reso}
\end{table}

\textbf{Standard Setting.} We first compare our attack with baselines in the standard setting, where we divide \texttt{CelebA} into private and public datasets. As shown in Table~\ref{table:1}, our method outperforms all baselines in terms of generative fidelity while exhibiting competitive attack accuracy across three models. The reconstructed images generated by our Diff-MI achieve the lowest FID and KNN Dist on all target classifiers. Specifically, we observe an average decrease of $20\%$ in FID and $50$ points in KNN Dist compared to the SOTA method (i.e., PLG-MI), indicating that our reconstruction is closely resembling the private images with improved generative fidelity and closer feature distance. We also note that our improvement in fidelity compromises attack accuracy compared to PLG-MI. Recall that our intention is to strive for superior accuracy-fidelity balance rather than solely pursuing high Acc. The improved fidelity is deemed more favorable when maintaining competitive Acc. Fig.~\ref{fig:3} visualizes the reconstructed images using different methods. Compared with KED-MI and PLG-MI, which are suffering from inferior generative fidelity because of GAN's inherent flaws and the fidelity degradation problem discussed in Sec.~\ref{sec:1}, our reconstruction exhibits more pronounced semantic facial features and demonstrates superior visual fidelity. We also compare the reconstruction using our method without fine-tuning (Ours w/o FT). The comparison indicates that our FT method effectively preserves the generative fidelity while improving attack accuracy.

\textbf{Distributional Shift Setting.} We further explore a more challenging yet realistic scenario where the public and private datasets exhibit larger distributional shifts. As shown in Table~\ref{table:2}, our Diff-MI outperforms all baselines in terms of generative fidelity as well and simultaneously achieves competitive attack accuracy against the SOTA method on both public datasets. It is evident that all methods suffer from a performance drop because of larger distributional shifts. Even so, our reconstruction still exhibits lower FID and KNN Dist with an average decrease of $20\%$ and an average reduction of $20$ points, respectively.

\begin{table*}[t]
\caption{\textbf{Diff-MI confronts MIA defense} (i.e., BiDO) and is compared with other baselines ($\mathcal{D}_{\text{pri}}$ = CelebA, $\mathcal{D}_{\text{pub}}$ = CelebA, Target Classifier = VGG16).}
\resizebox{\hsize}{!}{
% \scriptsize  % 调字体大小
\renewcommand{\arraystretch}{1.10}  % 调行距
\begin{tabular}{cccccccccccccc}
\toprule
                         & \textbf{Acc1 ↑}  & \textbf{Acc5 ↑}  & \textbf{FID ↓} & \textbf{PSNR ↑} & \textbf{SSIM ↑} & \textbf{FSIM ↑} & \textbf{VSI ↑} & \textbf{HaarPSI ↑} & \textbf{SR-SIM ↑} & \textbf{DSS ↑} & \textbf{MDSI ↓} & \textbf{LPIPS-A ↓} & \textbf{LPIPS-V ↓} \\ \midrule
GMI             & 23.40\%          & 47.07\%          & 28.04          & 13.09           & 0.39            & 0.69            & 0.86           & 0.40               & 0.82              & 0.44           & 0.57            & 0.3393             & 0.5245             \\
KED-MI          & 63.13\%          & 88.33\%          & 30.49          & 14.13           & 0.42            & 0.71            & 0.86           & 0.42               & 0.84              & 0.51           & 0.56            & 0.3324             & 0.5175             \\
PLG-MI          & \textbf{97.47\%} & \textbf{99.47\%} & 33.27          & 14.79           & 0.47            & 0.73            & 0.88           & 0.45               & 0.85              & 0.52           & 0.54            & 0.3025             & 0.4787             \\ \midrule
\textbf{Ours}            & 93.47\%          & 99.20\%          & \textbf{23.82} & \textbf{15.64}  & \textbf{0.53}   & \textbf{0.75}   & \textbf{0.89}  & \textbf{0.49}      & \textbf{0.87}     & \textbf{0.60}  & \textbf{0.53}   & \textbf{0.2998}    & \textbf{0.4747}    \\ \rowcolor[HTML]{dadada}
\textbf{Ours w/ defense} & 81.00\%          & 94.87\%          & 27.58          & 15.35           & 0.49            & 0.74            & 0.88           & 0.47               & 0.86              & 0.58           & \textbf{0.53}   & 0.3043             & 0.4825             \\ \bottomrule
\end{tabular}
}
\label{table:resconstruction_defense_Appendix}
\end{table*}

\begin{table}[t]
\caption{\textbf{Human preference study regarding fidelity and accuracy} ($\mathcal{D}_{\text{pri}}$ = CelebA, $\mathcal{D}_{\text{pub}}$ = CelebA, Target Classifier = VGG16). Results over $50\%$ indicate that workers prefer our reconstruction.} 
\centering
% \resizebox{\hsize}{!}{
\footnotesize % 调字体大小
\setlength\tabcolsep{5pt}  % 调列距
\begin{tabular}{c|cccc}
\toprule
\textbf{Ours} \textit{v.s.} & GMI \cite{zhang2020secret}     & KED-MI \cite{chen2021knowledge}     & PLG-MI \cite{yuan2023pseudo}     & Ours (w/o FT) \\ \midrule
\textbf{Fidelity}           & 68.97\%                        & 88.80\%                             & 85.92\%                          & 33.22\%       \\
\textbf{Accuracy}           & 68.92\%                        & 78.37\%                             & 73.44\%                          & 47.18\%       \\ \bottomrule
\end{tabular}
% }
\label{table:human}
\end{table}

\textbf{Reconstruction Quality.} In addition to attack accuracy (Acc) and distributional distance (FID and KNN Dist), we also compare the image quality of different reconstructions to provide a more comprehensive assessment of attack performance. As shown in Table~\ref{table:resconstruction_assessment}, our reconstructions outperform all baselines across all metrics in both settings. This indicates that our reconstruction is not merely a simple numerical approximation to the target classifier with high attack accuracy. Instead, it involves simultaneous exploration of private data at both pixel and feature levels, ensuring both accurate reconstruction and superior image quality. These metrics assess the similarity between reconstructed images and their corresponding private counterparts from various feature perspectives. This indicates that, in terms of reconstruction quality, our method is superior in both accuracy and fidelity. Whether at the pixel level or feature level, our method can more accurately and robustly reconstruct private features. This result also corroborates our discussion in Sec.~\ref{sec:discussion_on_metrics_appendix}. Although our method falls short in attack accuracy (Acc) compared with PLG-MI, it excels in terms of image similarity and quality (FID, KNN Dist, and PSNR/SSIM/LPIPS). This suggests that our method is not simply numerical fitting the target classifier but rather reconstructing privacy features more effectively from a human observation.

\textbf{Comparisons with VMI and PPA.} PPA introduces a plug-and-play MIA method and adopts a GAN pretrained on the public dataset. It intentionally relaxes the dependency between the target model and image prior, enabling the use of a single GAN to attack a wide range of target models. As shown in Table~\ref{table:PPA_appendix}, PPA suffers from low attack accuracy (Acc) and diminished reconstruction similarity (PSNR, SSIM, FSIM, etc.), even though it exhibits a decent FID. We argue that such an attack is suboptimal, as its lack of precision impedes the accurate reconstruction of user privacy, even coupled with respectable generative fidelity. This can be attributed to its plug-and-play attack paradigm, as its emphasis on attack convenience inherently affects the attack performance when applied to different target classifiers. In contrast, our Diff-MI purposely learns a target-specific generator (i.e., CDM) to approximate the target knowledge, resulting in improved and robust attack performance. Although training such a target-specific generator comes with additional costs, we believe that these sacrifices are worthwhile in robustly achieving superior attack performance. VMI formulates the MIA problem as a variational inference problem, and provides a framework using deep normalizing flows \cite{kingma2018glow}. This framework alleviates the fidelity degradation problem by incorporating a regularization term (i.e., KL divergence) into the distribution of GAN's latent space. As shown in Table~\ref{table:VMI_appendix}, our Diff-MI outperforms VMI across nearly all metrics. This superiority can be attributed to our proposed target-specific CDM, which not only ensures generative fidelity but also achieves more precise reconstruction.

\textbf{High Resolution.} We conduct our Diff-MI in high-resolution scenarios following PPA's \cite{struppek2022plug} setup. We follow the same experimental setup as PPA \cite{struppek2022plug}, the first work to extend MIAs to high-resolution scenarios. Specifically, we introduce CelebA as $\mathcal{D}_{\text{pri}}$ and align images using the HD CelebA Cropper \cite{HD-CelebA-Cropper}. We crop images using a face factor of 0.65 and resize them to $224 \times 224$ with bicubic interpolation (other parameters are set as default). Then following PPA, we take the first 1,000 classes with the most number of samples out of total 10,177 classes to train the target classifier $T$ = IR152 \cite{he2016deep}. Meanwhile, we use the same training data ($\mathcal{D}_{\text{pri}}$ = CelebA) to train an Inception-v3 model \cite{szegedy2016rethinking} as the evaluation classifier. Training details are the same as PPA without any modification using their official code. During the attack stage, we re-implement PPA following its default configuration and reconstruct 50 images for the first 100 target classes for evaluation. As for our Diff-MI, we introduce LDM \cite{rombach2022high}, which can encode input images from $256 \times 256$ into $64 \times 64$ for computational efficiency, to train the target-specific CDM with $\mathcal{D}_{\text{pub}}$ = FFHQ in the latent space. As shown in Table~\ref{table:high_reso}, our method can achieve better attack accuracy with competitive generative fidelity. This can be attributed to our target-specific generator compared to PPA's utilization of pretrained StyleGAN2 \cite{karras2020analyzing}. Although training such a target-specific generator comes with additional costs, we believe these sacrifices are worthwhile in robustly achieving superior attack performance.

\textbf{MIAs with Defenses.} To defend against MIAs, \cite{wang2021improving} has proposed to employ a unilateral dependency optimization strategy to minimize the dependency between inputs and outputs during training the target model. Subsequently, \cite{peng2022bilateral} enhanced this approach by introducing a bilateral dependency optimization (BiDO) strategy and achieved SOTA performance in defending MIAs. The BiDO strategy aims to minimize the dependency between the latent representations and the inputs while maximizing the dependency between latent representations and the outputs. Hence, we conduct an experiment to assess the attack performance of our Diff-MI with defenses (i.e., BiDO), particularly focusing on its performance in terms of generative quality. As shown in Table~\ref{table:resconstruction_defense_Appendix}, there is an evident decrease in Acc (i.e., $93.47\% \to 81.00\%$), along with a slight decline in the reconstruction quality (e.g., FID: $23.82 \to 27.58$). Even so, our reconstruction quality still surpasses that of other baselines (i.e., FID, PSNR, SSIM, FSIM, etc.), demonstrating the robustness of our Diff-MI in reconstructing high-quality private images.

\begin{figure*}[t]
    \centering
    \includegraphics[width=\hsize]{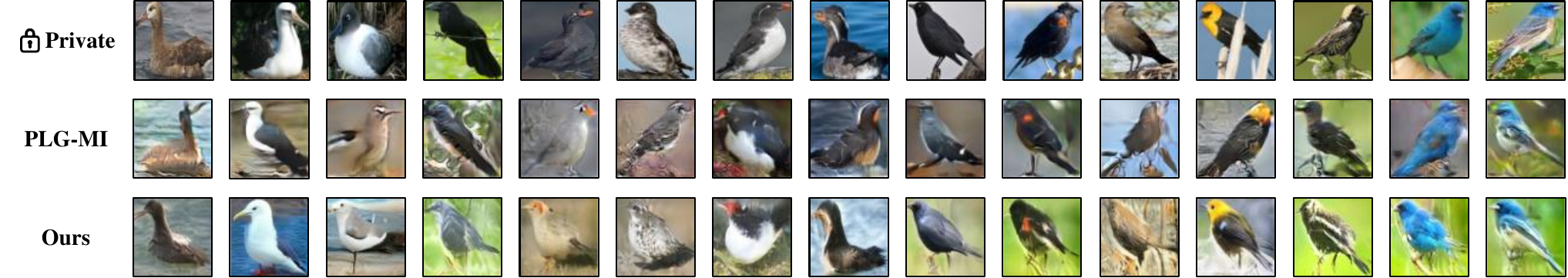}
    \caption{Visual comparison of reconstructed images labeled from ``001. Black-footed Albatross" to ``015. Lazuli Bunting" between PLG-MI and our Diff-MI on CUB-200-2011.}
    \label{fig:3_appendix}
\end{figure*}

\begin{figure}[t]
    \centering
    \includegraphics[width=\hsize]{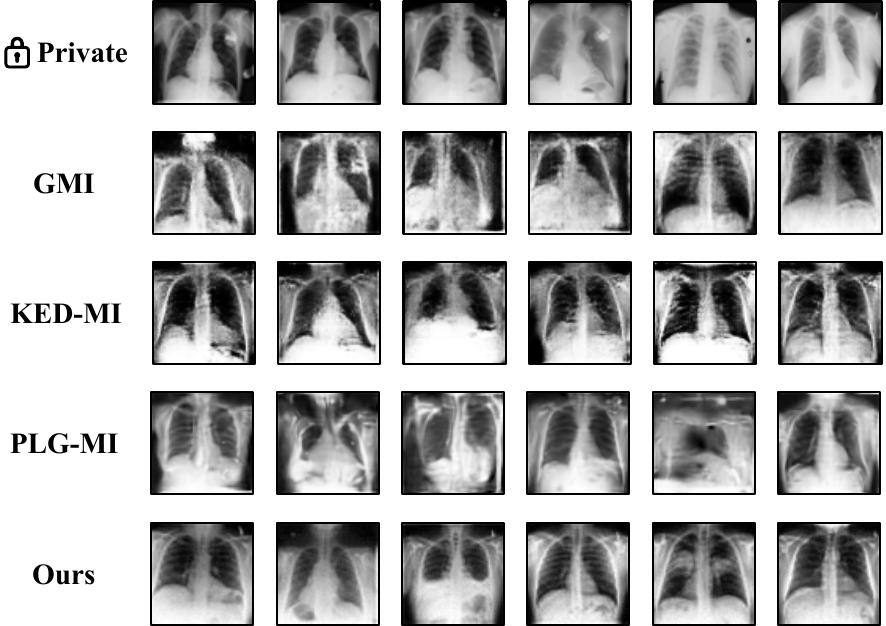}
    \caption{Visual comparison of reconstructed chest X-rays labeled from ``0" to ``5" using different MIA methods ($\mathcal{D}_{\text{pri}}$ = ChestX-Ray, $\mathcal{D}_{\text{pub}}$ = CheXpert).}
    \label{fig:4_appendix}
\end{figure}

\begin{table}[t]
\caption{\textbf{Attack performance comparison on CUB-200-2011 and ChestX-Ray.} ``$A \to B$" represents the diffusion model and target classifier trained on datasets $A$ and $B$, respectively.}
\centering
% \resizebox{\hsize}{!}{
\footnotesize  % 调字体大小
\renewcommand{\arraystretch}{1.10}  % 调行距
\setlength\tabcolsep{4pt}  % 调列距
\begin{tabular}{cccccc}
\toprule
                                                                                                        & Method    & \textbf{Acc1 ↑}  & \textbf{Acc5 ↑}  & \textbf{FID ↓} & \textbf{KNN Dist ↓} \\ \midrule
\multirow{4}{*}{\textbf{\begin{tabular}[c]{@{}c@{}}CUB-200-2011\\      ↓\\  CUB-200-2011\end{tabular}}} & GMI       & 20.00\%         & 49.40\%         & 167.06         & 954.04          \\
                                                                                                        & KED-MI    & 46.20\%         & 78.60\%         & 199.51         & 912.02              \\
                                                                                                        & PLG-MI    & 70.60\%         & 90.60\%         & 116.91         & 915.00    \\
                                                                                                        & \cellcolor[HTML]{dadada}\textbf{Ours}   & \cellcolor[HTML]{dadada}\textbf{79.20\%}    & \cellcolor[HTML]{dadada}\textbf{94.00\%}    & \cellcolor[HTML]{dadada}\textbf{58.06}    & \cellcolor[HTML]{dadada}\textbf{903.05} \\ \midrule
\multirow{4}{*}{\textbf{\begin{tabular}[c]{@{}c@{}}ChestX-Ray\\      ↓\\      ChestX-Ray\end{tabular}}} & GMI    & 13.21\%          & 51.86\%          & 127.06         & 117.77              \\
                                                                                                        & KED-MI & 52.86\%          & 91.07\%          & 100.74         & 128.57              \\
                                                                                                        & PLG-MI & 37.43\%          & 76.14\%          & 150.41         & \textbf{107.37}     \\
                                                                                                        & \cellcolor[HTML]{dadada}\textbf{Ours}   & \cellcolor[HTML]{dadada}\textbf{63.50\%} & \cellcolor[HTML]{dadada}\textbf{90.93\%} & \cellcolor[HTML]{dadada}\textbf{69.24} & \cellcolor[HTML]{dadada}108.54              \\ \midrule
\multirow{4}{*}{\textbf{\begin{tabular}[c]{@{}c@{}}CheXpert\\      ↓\\      ChestX-Ray\end{tabular}}}   & GMI    & 14.29\%          & 46.50\%          & 214.46         & 139.30              \\
                                                                                                        & KED-MI & 50.50\%          & 85.86\%          & 153.79         & \textbf{124.52}     \\
                                                                                                        & PLG-MI & 34.00\%          & 80.00\%          & 119.80         & 168.03              \\
                                                                                                        & \cellcolor[HTML]{dadada}\textbf{Ours}   & \cellcolor[HTML]{dadada}\textbf{70.36\%} & \cellcolor[HTML]{dadada}\textbf{92.43\%} & \cellcolor[HTML]{dadada}\textbf{74.40} & \cellcolor[HTML]{dadada}126.77              \\ \bottomrule
\end{tabular}
% }
\label{table:3_appendix}
\end{table}

\begin{figure*}[t]
    \centering
    \subfloat[]{\includegraphics[width=.24\hsize]{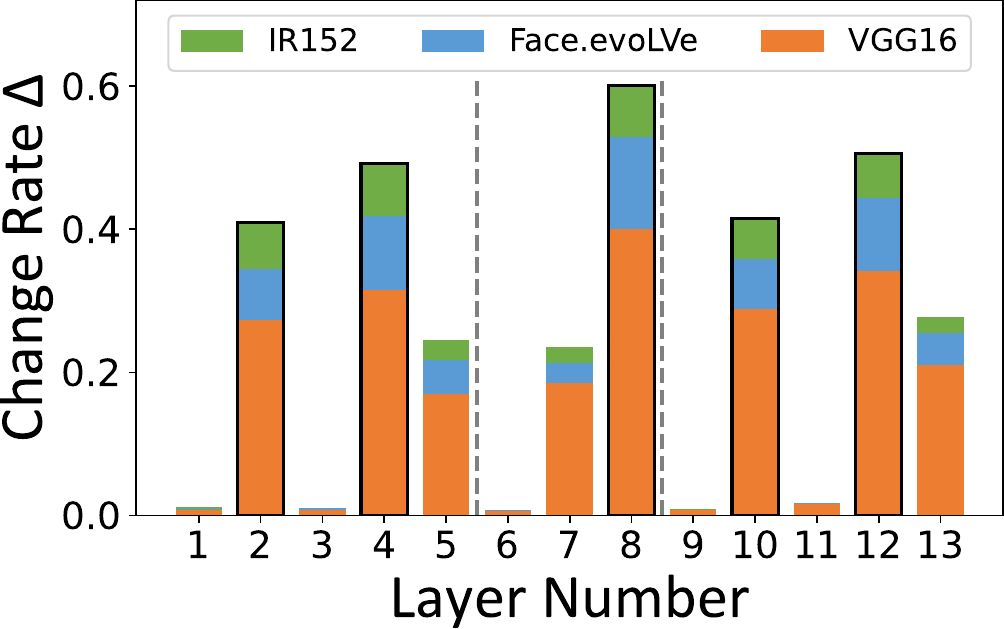}}\hspace{2pt}
    \subfloat[]{\includegraphics[width=.24\hsize]{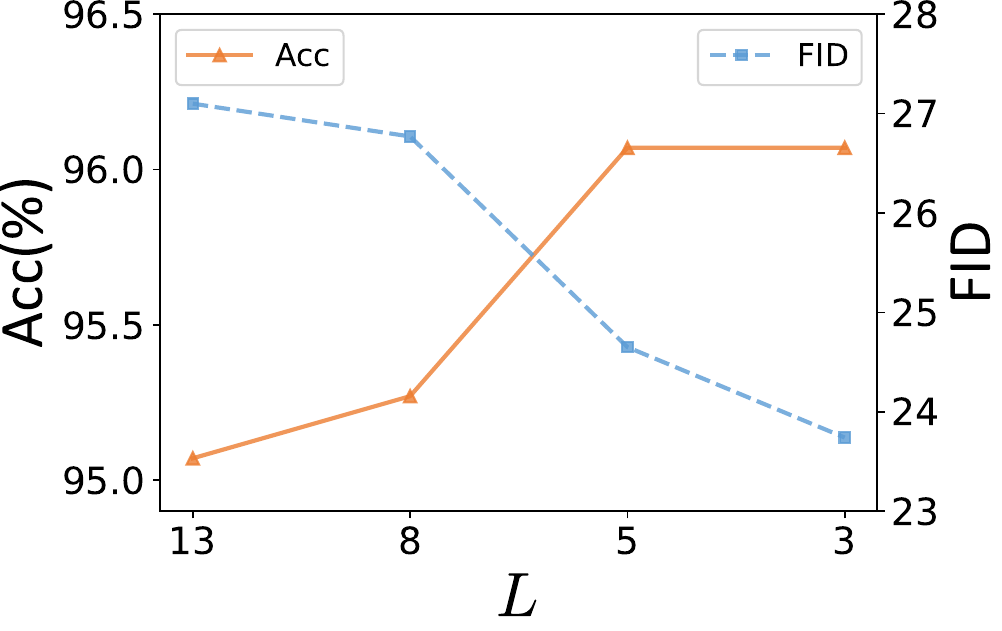}}\hspace{2pt}
    \subfloat[]{\includegraphics[width=.24\hsize]{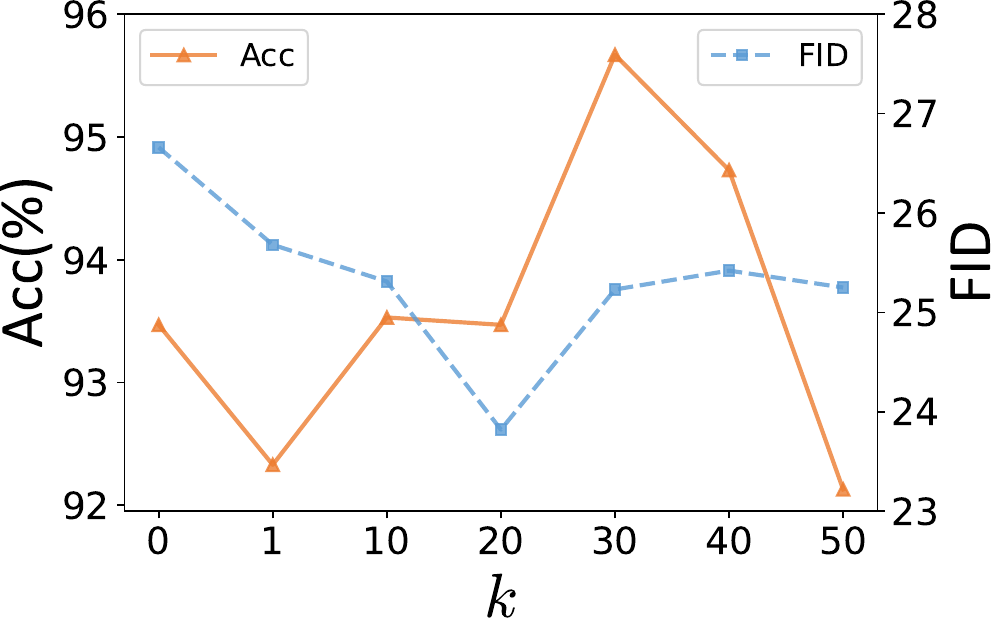}}\hspace{2pt}
    \subfloat[]{\includegraphics[width=.24\hsize]{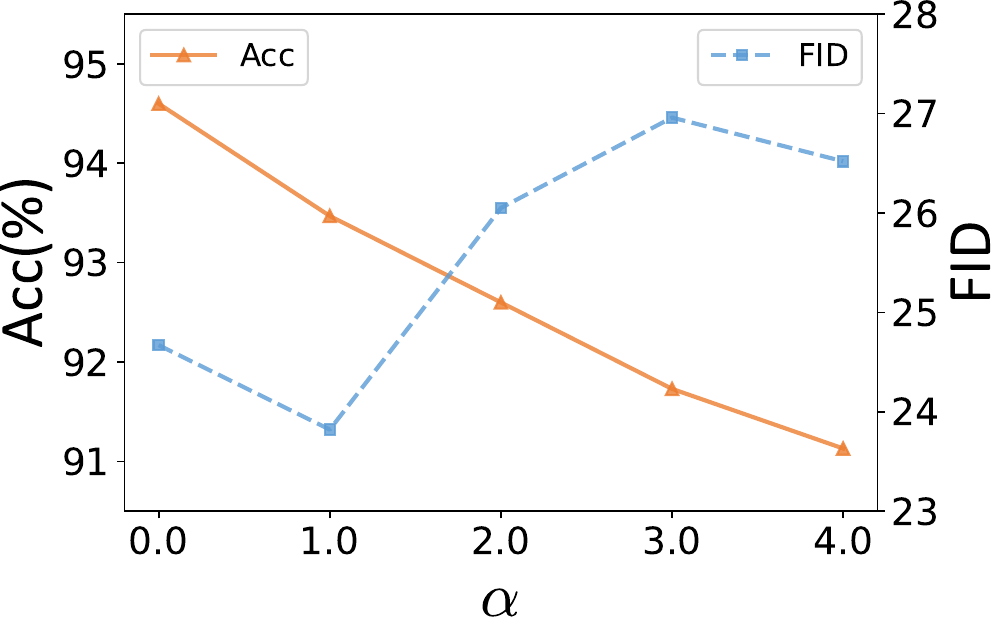}}
    \caption{
    (a) Analysis of change rates in the middle block. We observe similar plots for three target classifiers. The middle block consists of one attention block (6 $\sim$ 8) and two residual blocks (1 $\sim$ 5 and 9 $\sim$ 13). Among all 13 layers in the middle block, the changes in certain layers (i.e., 2, 4, 8, 10, 12) are relatively higher than the others, indicating that they play a significant role during fine-tuning.
    (b) Acc and FID with fine-tuning the label embedding layer and different layers in the middle block, where $L$ means fine-tuning the top-$L$ layers that exhibit the highest $\Delta$.
    (c) and (d) compare the attack performance of different hyperparameters $k$ and $\alpha$ in $\mathcal{L}_{\text{cls}}$.
    (a) - (d) follow the same experimental setup: $\mathcal{D}_{\text{pri}}$ = CelebA, $\mathcal{D}_{\text{pub}}$ = CelebA, and for (b) - (d): $T$ = VGG16.
    }
    \label{fig:4}
\end{figure*}

\subsection{Comparisons on Other Datasets}

In addition to the face recognition task, we also conduct MIAs on other classification tasks, including fine-grained image classification on \texttt{CUB-200-2011} \cite{wah2011caltech} and disease prediction on \texttt{ChestX-Ray} \cite{wang2017chestx} and \texttt{CheXpert} \cite{irvin2019chexpert}.

\textbf{Experimental Details.} For \texttt{CUB-200-2011}, we use the officially provided bounding-box annotations to crop each sample in advance. Among all 200 classes, we designate the first 50 classes containing 2,889 images as the private dataset to train the target classifier ResNet-18 \cite{he2016deep}, while the remaining 150 classes containing 8,899 images as the public dataset to train the evaluation model ResNet-34 \cite{he2016deep}.
For \texttt{ChestX-Ray}, we randomly select 2,000 images from each of the 14 classes. If the total number of images in a particular class is less than 2,000, we include all the images of that class, resulting in a total of 25,344 images as the private dataset. We adopt ResNet-18 \cite{he2016deep} trained on the private dataset as the target model. Then, we randomly select 20,000 images from the class with the label ``NoFinding" as the public dataset to train the evaluation model ResNet-34 \cite{he2016deep}. Furthermore, we also introduce another large-scale dataset of chest X-rays 
\texttt{CheXpert} \cite{irvin2019chexpert}, consisting of 224,316 chest radiographs of 65,240 patients. We select 12,940 images with equal height and width labeled as ``view1" as the public dataset. All images from the above datasets are resized to $64 \times 64$. $n$ for the top-$n$ selection strategy and $k$ in the top-$k$ loss are set to $200$ and $5$ for \texttt{CUB-200-2011} and $1000$ and $3$ for \texttt{ChestX-Ray}. $\alpha$ for the p-reg loss is set to $1.0$ as well.
In step-1, we first pretrain the CDM with batch size $150$ for 50K iterations on two A40 GPUs and then fine-tune the pretrained CDM for all target classes, conducting a total of $400$ epochs for both \texttt{CUB-200-2011} and for \texttt{ChestX-Ray}. In step-2, we follow the same setup of the face recognition task and reconstruct $10$ and $100$ images per class for \texttt{CUB-200-2011} and \texttt{ChestX-Ray} for quantitative evaluation.

\begin{table}[t]
\caption{\textbf{Ablation study} of the various components in Diff-MI ($\mathcal{D}_{\text{pri}}$ = CelebA, $\mathcal{D}_{\text{pub}}$ = CelebA, $T$ = VGG16). \ding{55}$_\textbf{1}$ and \ding{55}$_\textbf{2}$ represent replacing our proposed  $\mathcal{L}_{\text{cls}}$ with CE loss and max-margin loss, respectively.}
% \resizebox{\hsize}{!}{
\footnotesize  % 调字体大小
\renewcommand{\arraystretch}{1.1}  % 调行距
\begin{tabular}{ccc|cccc}
    \toprule
    \multicolumn{3}{c|}{\textbf{Method}}           & \multirow{2}{*}[-2.5pt]{\textbf{Acc1 ↑}} & \multirow{2}{*}[-2.5pt]{\textbf{Acc5 ↑}} & \multirow{2}{*}[-2.5pt]{\textbf{FID ↓}} & \multirow{2}{*}[-2.5pt]{\textbf{KNN Dist ↓}} \\ \cmidrule(r){1-3}
    FT      & PGD     & $\mathcal{L}_\text{cls}$                                                                                                                                                      \\ \midrule 
    \ding{51} & \ding{51} & \ding{51}                    & \textbf{93.47\%}                 & \textbf{99.20\%}                 & 23.82                           & \textbf{1081.98}                     \\
    \ding{51} & \ding{55} & \ding{51}                    & 90.47\%                          & 98.07\%                          & 24.33                           & 1107.16                              \\
    \ding{51} & \ding{51} & \ding{55}$_\textbf{1}$         & 88.73\%                          & 98.13\%                          & 24.98                           & 1100.37                              \\
    \ding{51} & \ding{51} & \ding{55}$_\textbf{2}$         & 91.20\%                          & 98.40\%                          & 25.06                           & 1102.28                              \\
    \ding{55} & \ding{51} & \ding{51}                    & 75.87\%                          & 93.60\%                          & \textbf{20.95}                  & 1116.79                              \\ \bottomrule
\end{tabular}
% }
\label{table:ablation}
\end{table}

\textbf{Experimental Results.} As shown in Table~\ref{table:3_appendix}, our method demonstrates superior performance across all evaluation metrics on \texttt{CUB-200-2011}. This indicates that quantitatively, our reconstruction outperforms all baselines in terms of both attack accuracy and generative fidelity, two critical factors in assessing the attack performance of MIAs. Our method achieves an impressive FID reduction of over 50\% compared to all baselines. This substantial improvement signifies our ability to generate images of exceptional fidelity that closely resemble real examples from the dataset. Notably, our Diff-MI, which is based on diffusion models, requires minimal hyperparameter tuning when generalizing to new datasets, while previous GAN-based baselines were highly sensitive to hyperparameter selection. This sensitivity necessitates additional hyperparameter tuning when attempting to generalize them to new datasets. The reconstructed images obtained by PLG-MI and our Diff-MI are visualized in Fig.~\ref{fig:3_appendix}. The visualization clearly demonstrates that our reconstruction effectively captures intricate details of various bird species, including beak shapes, wing coloration, and tail length. Whereas PLG-MI exhibits a ringing-like effect due to its sensitivity to hyperparameters, resulting in a degradation of generative fidelity. Conclusively, our method not only achieves superior generative fidelity but also accurately captures the semantic information of private images.

Table~\ref{table:3_appendix} presents the results of performing MIAs on the private medical dataset \texttt{ChestX-Ray}, taking \texttt{ChestX-Ray} and \texttt{CheXpert} as the public dataset, respectively. Compared to all baselines, our method achieves nearly the best performance across all metrics, with only a slight increase in KNN Dist. Particularly noteworthy is our substantial improvement in generative fidelity, as evidenced by an average decrease of $45\%$ in FID. This significant reduction in FID highlights the enhanced realism and fidelity of the reconstructed samples, indicating the superiority of our Diff-MI over SOTA methods in MIAs. These compelling results pave the way for broader applications of our method in various domains where high-quality MIAs are essential. We also provide a visual comparison of the private images reconstructed by baselines and our method using \texttt{CheXpert} as the public dataset, as shown in Fig.~\ref{fig:4_appendix}.

\subsection{Human Preference Study}

We conduct a human preference study using Amazon Mechanical Turk. We perform a paired test of our method with three baselines and Ours (w/o FT). For generative fidelity, we display two reconstructed images from each method (ours v.s. baseline) with the question - \textit{``Which image do you consider to be more realistic?".} For attack accuracy, we display three target images of the same class from $\mathcal{D}_{\text{pri}}$, along with the paired reconstructed images, and ask - \textit{``Which image do you consider to be more similar to the target images?".} As shown in Table~\ref{table:human}, our reconstruction is preferred ($\geq 50\%$) over all baselines in terms of both generative fidelity and attack accuracy. Workers inevitably lean to images with higher fidelity when discerning more similar images. Therefore, compared with Ours (w/o FT), our method falls short in both aspects. Attackers can choose whether to fine-tune depending on superior human preference or better attack performance.

\subsection{Ablation Study} \label{sec:ablation_study}

We first compare the impact of fine-tuning different layers in the middle block, considering various change rates. Results in Fig.~\ref{fig:4} (b) indicate that fine-tuning fewer layers yields superior attack performance. Specifically, among all layers in the middle block, only fine-tuning the top-$3$ layers achieves the highest Acc and lowest FID but suffers from a long time for convergence. Thus we choose to fine-tune the top-$5$ layers in our implementation. Fig.~\ref{fig:4} (c) shows the top-$k$ loss improves the attack performance and $k=20$ is optimal for the lowest FID. Increasing $k$ (e.g. $30$, $40$) leads to higher Acc but inferior fidelity proved by the increasing FID, meaning that the reconstruction starts to incorporate some semantically meaningless features that overfit the target classifier. Fig.~\ref{fig:4} (d) illustrates the p-reg loss serves as a beneficial regularization term for improving generative fidelity ($\alpha=1.0$), albeit with a slight decrease in Acc. Conclusively, $k$ and $\alpha$ are set to $20$ and $1.0$, striking a trade-off between Acc and FID. 

We further ablate certain key components in Diff-MI and derive the following conclusions from Table~\ref{table:ablation}: (1) The PGD method significantly enhances attack performance in terms of both attack accuracy and generative fidelity, owing to its comprehensive utilization of the target classifier; (2)  Our improved loss function demonstrates superior performance compared to the CE loss and max-margin loss employed in previous studies. This indicates that our method can accurately reconstruct more discriminative images in MIAs; (3) Our target-specific CDM effectively captures the knowledge of the target classifier, resulting in higher Acc and closer feature distance. However, it does encounter a trade-off between generative fidelity and attack accuracy during the fine-tuning process. Attackers can choose whether to fine-tune the pretrained generator based on their specific requirements, prioritizing either higher attack accuracy or improved generative fidelity.

\section{Conclusion}

This paper has presented a two-step diffusion-based white-box MIA method, namely Diff-MI. In step-1, we customize a target-specific CDM to distill the target classifier’s knowledge and approximate its private distribution under a pretrain-then-finetune fashion, achieving superior accuracy-fidelity balance. In step-2, we introduce an iterative image reconstruction method to involve both diffusion priors and target classifier’s knowledge through a combinatorial optimization problem. Additionally, we introduce an improved max-margin loss for MIAs, optimizing the use of feature information and soft labels from the target classifier with better attack performance. Experiments show that our method achieves SOTA generative fidelity and reconstruction quality with competitive attack accuracy in diverse scenarios across different model architectures.

\textbf{Limitation.} To enhance the robustness and overall attack performance in MIAs, our method requires training a target-specific CDM from scratch, incurring more computational costs. Additionally, our Diff-MI still suffers from a trade-off between generative fidelity and attack accuracy during fine-tuning and attacking. For future work, we intend to explore the potential of pretrained diffusion models with enhanced generative quality and attack accuracy in both white-box and black-box MIAs.

\textbf{Ethical Statement.} MIAs can have profound impacts on society as they pose a serious threat to individual privacy and data security, eroding public trust in artificial intelligence systems and hindering technology adoption in critical applications. Meanwhile, diffusion models are gaining popularity in the field of image synthesis, whereas they can also be employed by malicious attackers for conducting generative MIAs and reconstructing high-quality private images about users. While our Diff-MI could potentially have negative societal impacts, we intend to reveal the vulnerabilities of current systems and raise public awareness about privacy attacks on DNNs. We believe that our work will inspire the design of defenses against MIAs.

\textbf{Data Availability.} In our experiments, we simulate the real-world attack scenario by introducing \texttt{CelebA}\footnote{\url{https://mmlab.ie.cuhk.edu.hk/projects/CelebA.html}} \cite{liu2015deep}, \texttt{FFHQ}\footnote{\url{https://github.com/NVlabs/ffhq-dataset}} \cite{karras2019style}, and \texttt{FaceScrub}\footnote{\url{https://vintage.winklerbros.net/facescrub.html}} \cite{ng2014data} as the private and public datasets to inverse face recognition classifiers. Meanwhile, we also compare other plausible scenarios, such as fine-grained image classification and disease detection, where \texttt{CUB-200-2011}\footnote{\url{https://www.vision.caltech.edu/datasets/cub_200_2011/}} \cite{wah2011caltech}, \texttt{ChestX-Ray}\footnote{\url{https://www.kaggle.com/datasets/nih-chest-xrays/data}} \cite{wang2017chestx}, and \texttt{CheXpert}\footnote{\url{https://aimi.stanford.edu/datasets/chexpert-chest-x-rays}} \cite{irvin2019chexpert} are introduced. More details about datasets, models, and codes of this paper are available in our GitHub repository\footnote{\url{https://github.com/Ouxiang-Li/Diff-MI}}.

\bibliographystyle{IEEEtran}
\bibliography{reference.bib}

\end{document}